\documentclass[a4paper,twoside]{article}

\usepackage{epsfig}
\usepackage{subcaption}
\usepackage{calc}
\usepackage{amssymb}
\usepackage{amstext}
\usepackage{amsmath}
\usepackage{amsthm}
\usepackage{multicol}
\usepackage{pslatex}

\usepackage{natbib}
\let\bibhang\relax
\usepackage{apalike}

\usepackage[bottom]{footmisc}
\usepackage[utf8]{inputenc}
\usepackage[table]{xcolor}
\usepackage{multirow}
\usepackage{enumitem}
\usepackage{tabularx}
\usepackage{ragged2e}
\usepackage[tableposition=top]{caption}
\usepackage{stfloats}

\usepackage[pagebackref=true,breaklinks=true,colorlinks,bookmarks=false]{hyperref}

\newcommand\Tstrut{\rule{0pt}{2.6ex}}         
\newcommand\Bstrut{\rule[-0.8ex]{0pt}{0pt}}

\usepackage{SCITEPRESS}     

\newcommand{\pps}{panoptic-part segmentation}
\newcommand{\PPS}{Panoptic-Part Segmentation}
\newcommand{\ours}{JPPF (Ours)}
\newcommand{\things}{\emph{things}}
\newcommand{\stuff}{\emph{stuff}}

\usepackage[sort&compress,capitalize,noabbrev,nameinlink]{cleveref}
\creflabelformat{equation}{#2#1#3}

\begin{document}

\title{Multi-task Fusion for Efficient \PPS}

\author{\authorname{Sravan Kumar Jagadeesh, René Schuster, and Didier Stricker}
\affiliation{DFKI -- German Research Center for Artificial Intelligence, Kaiserslautern, Germany}
\email{firstname.lastname@dfki.de}
}

\abstract{In this paper, we introduce a novel network that generates semantic, instance, and part segmentation using a shared encoder and effectively fuses them to achieve \pps. Unifying these three segmentation problems allows for mutually improved and consistent representation learning. To fuse the predictions of all three heads efficiently, we introduce a parameter-free joint fusion module that dynamically balances the logits and fuses them to create \pps. Our method is evaluated on the Cityscapes Panoptic Parts (CPP) and Pascal Panoptic Parts (PPP) datasets. For CPP, the $PartPQ$ of our proposed model with joint fusion surpasses the previous state-of-the-art by 1.6 and 4.7 percentage points for all areas and segments with parts, respectively. On PPP, our joint fusion outperforms a model using the previous top-down merging strategy by 3.3 percentage points in $PartPQ$ and 10.5 percentage points in $PartPQ$ for partitionable classes.}


\onecolumn \maketitle \normalsize \setcounter{footnote}{0} \vfill

\section{\uppercase{Introduction}}
\label{sec:intro}

\begin{figure}[t]
    \centering
    \begin{tabularx}{\linewidth}{m{2.5mm} X}
    \rotatebox{90}{\textbf{Image}}        &   \includegraphics[width=\linewidth]{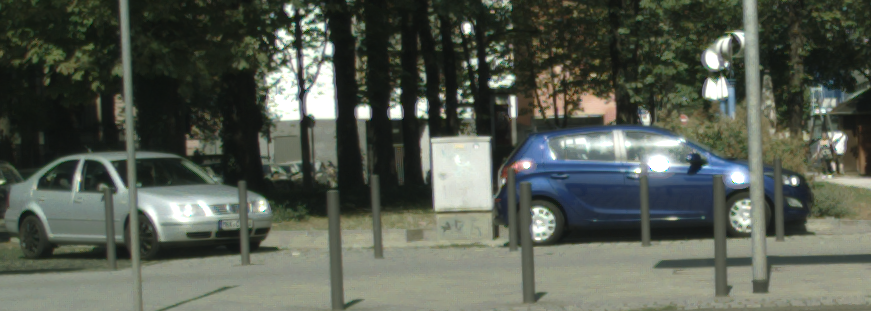}\\%
    \rotatebox{90}{\textbf{Ground-truth}}        &   \includegraphics[width=\linewidth]{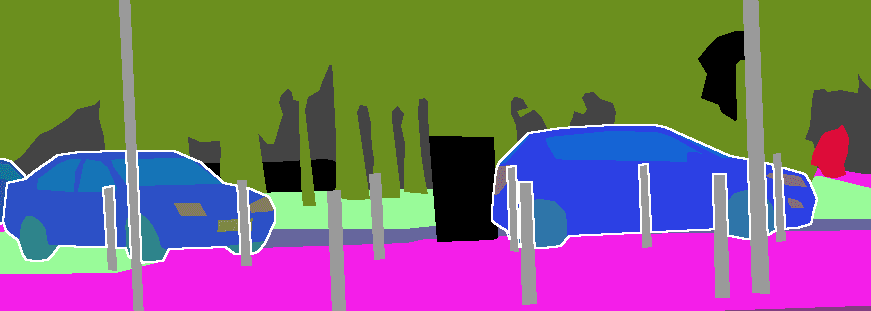}\\%
    \rotatebox{90}{\textbf{Baseline$^{\ast}$}}        &   \includegraphics[width=\linewidth]{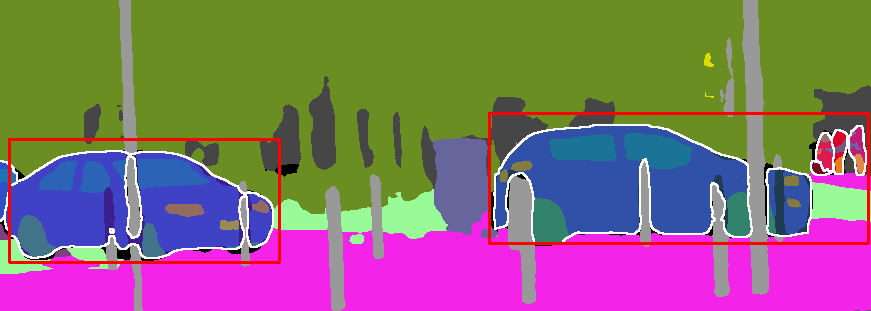}\\%
    \rotatebox{90}{\textbf{\ours{}}}        &   \includegraphics[width=\linewidth]{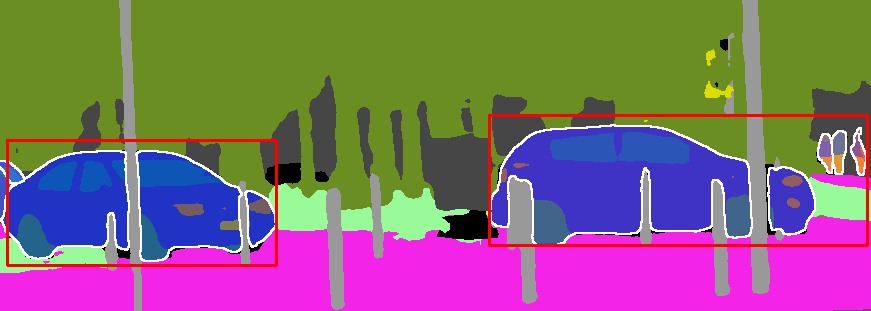}\\%
    \end{tabularx}
    \caption{We propose a unified network with Joint Panoptic Part Fusion (JPPF) to generate \pps. Here, a prediction of our proposed model on CPP \citep{meletis2020cityscapes} is shown. Details about the baseline are given in \cref{sec:results:sota}.}
    \label{fig:teaser}
\end{figure}

The human eye can observe a scene at various levels of abstraction.
Humans can not only view the scene and differentiate semantic categories such as bus, car, and sky, but they can also understand them.
However, they can also distinguish between the parts of each entity, such as car windows and bus chassis, and group them according to their instances.
There is no deep learning approach that seeks to achieve several layers of abstraction with a single network at the moment.

The two pieces that make up a scene are \stuff{} \ and \things{} \citep{cordts2016cityscapes}.
Things are countable amorphous objects such as persons, cars, or buses, whereas \stuff{} like the sky or road is usually not countable.
Many tasks have been created to identify these aspects in an image. Semantic segmentation and instance segmentation are two of the most common tasks.
However, these methods are incapable of describing the entire image. Scene parsing was created to fill this void, with the goal of describing the entire image by recognizing and semantically segmenting both \stuff \ and \things, a process which is known as panoptic segmentation \citep{PanopticSegKiri}.
This approach has introduced several state-of-the-art panoptic segmentation methods \citep{ChengPD2019,kirillov2019panoptic,li2020unifying,mohan2021efficientps,PorziSSS,yuwenUPSnet}.
Part segmentation, or part parsing, on the other hand, seeks to semantically analyze the image based on part-level semantics for each class.
There has been some effort in this area, but often part segmentation has been treated as a semantic segmentation problem \citep{gong2019graphonomy,jiang2018cnn,jiang2019cnn,li2017holistic,liu2018cross,luo2013pedestrian}.
There are a few instance-aware methods \citep{gong2018instance,li2017holistic,zhao2018understanding} and even fewer that handle multi-class part objects \citep{zhao2019multi,michieli2020gmnet}.

Part-aware panoptic segmentation \citep{de2021part,li2022panoptic} was recently introduced to unify semantic, instance, and part segmentation. 
An example of part-aware panoptic segmentation is shown in \cref{fig:teaser}.
In \citep{de2021part}, a baseline approach is presented in which two networks are used, one for panoptic segmentation and the other for part segmentation. These two networks are trained independently and the results of both are combined using a uni-directional (top-down) merging strategy.
This technique of independent training has significant drawbacks.
Due to the use of two different networks, there is a computational overhead.
As the authors employ different networks, there will be no consistency in their predictions, making the merging process inefficient.
Also, the independent training strategy leads to learning redundancy since they could potentially share semantic information between segmentation heads.

In this work, we propose a joint network that uses a shared feature extractor to perform semantic, instance, and part segmentation. To achieve \pps, we propose Joint Panoptic Part Fusion (JPPF), which fuses all three predictions by giving equal priority to each prediction head.
The following is a summary of key contributions of this paper:
 
 \begin{itemize}
    \item We present a single new network that uses a shared encoder to perform semantic, instance, and part segmentation and fuse them efficiently to produce \pps.
    \item To achieve \pps, we propose a parameter-free joint panoptic part fusion module that dynamically considers the logits from the semantic, instance, and part head and consistently integrates the three predictions.
    \item We conduct a thorough analysis of our approach and demonstrate the shared encoder's efficacy and the consistency of the novel, joint fusion strategy.
    \item When compared to state-of-the-art \citep{de2021part}, our suggested fusion yields denser results at a higher quality.
 \end{itemize}

\section{\uppercase{Related Work}}
Part-aware panoptic segmentation \citep{de2021part} is a recently introduced problem that brings semantic, instance, and part segmentation together. There have been several methods proposed for these individual tasks, including panoptic segmentation, which is a blend of semantic and instance segmentation.

\subsection{Towards \PPS{}} \label{sec:related:segmentation}

\paragraph{Semantic Segmentation.}
PSPnet \citep{zhao2017pyramid} introduced the pyramid pooling module, which focuses on the importance of multi-scale features by learning them at many scales, then concatenating and up-sampling them. \citet{ChenPSA17} proposed Atrous Spatial Pyramid Pooling (ASPP), which is based on spatial pyramid pooling and combines features from several parallel atrous convolutions with varying dilation rates, as well as global average pooling. The incorporation of multi-scale characteristics and the capturing of global context increases computational complexity. So, \citet{chen2018a} introduced the Dense Predtiction Cell (DPC)  and \citet{Valada2018} suggested multi-scale residual units with changing dilation rates to compute high-resolution features at various spatial densities, as well as an efficient atrous spatial pyramid pooling module called eASPP to learn multi-scale representation with fewer parameters and a broader receptive field. In the encoder-decoder architecture, a lot of effort has been advocated for improving the decoder's upsampling layer.
\citet{chen2018b} extend DeepLabV3 \citep{ChenPSA17} by adding an efficient decoder module to enhance segmentation results at object boundaries. Later, \citet{Tian2019} suggest replacing it with data-dependent up-sampling (DUpsampling), which can recover pixel-wise prediction from low-resolution CNN outputs and take advantage of the redundant label space in semantic segmentation. 

\paragraph{Instance Segmentation.}
Here, we mainly concentrate on proposal based approaches.
\citet{HariharanAGM14} proposed a simultaneous object recognition and segmentation technique that uses Multi-scale Combinatorial Grouping (MCG) \citep{Pont-TusetABMM15} to generate proposals and then run them through a CNN for feature extraction.
In addition, \citet{HariharanAGM14a} presented a hyper-column pixel descriptor that captures feature representations of all layers in a CNN with a strong correlation for simultaneous object detection and segmentation.
\citet{PinheiroCD15} proposed the DeepMask network, which employs a CNN to predict the segmentation mask of each object as well as the likelihood of the object being in the patch. FCIS \citep{LiFCIS2017} employs position sensitive inside/outside score maps to simultaneously predict object detection and segmentation.  Later, one of the most popular networks for instance segmentation, Mask-RCNN \citep{HeGDG17}, was introduced. It extends Faster-RCNN \citep{RenHG015} with an extra network that segments each of the detected objects. RoI-align, which preserves exact spatial position, replaces RoI-pool, which performs coarse spatial quantization for feature encoding.

\paragraph{Part Segmentation.}
Dense part level segmentation, on the other hand, is instance agnostic and is regarded as a semantic segmentation problem \citep{gong2019graphonomy,jiang2018cnn,jiang2019cnn,li2017holistic,liu2018cross,luo2018trusted,michieli2020gmnet,zhao2019multi}. Most of the research has been conducted to perform human part parsing \citep{zhao2018understanding,gong2018instance,dong2013deformable,ladicky2013human,li2020self,liang2018look,lin2020cross,ruan2019devil,yang2019parsing}, and only little work has addressed multi-part segmentation tasks \citep{zhao2019multi, michieli2020gmnet}.

\paragraph{Panoptic Segmentation.}
The authors of \citep{PanopticSegKiri} combined the output of two independent networks for semantic and instance segmentation and coined the term panoptic segmentation. Panoptic segmentation approaches can be divided into top-down methods \citep{LiAnurag2018, LiuPS2019, Jie2018, yuwenUPSnet, SofibarkonAdaptIS, PorziSSS} that prioritize semantic segmentation prediction and bottom-up methods \citep{Tien2019DL, ChengPD2019, NaiyuSSAP2019} that prioritize instance prediction. In this work, we build on EfficientPS \citep{mohan2021efficientps} which will be extended to perform \pps.

\subsection{\PPS{}.} \label{sec:related:partpanoptic}
In recent years, Part-Aware Panoptic Segmentation \citep{de2021part} was introduced, which aims at a unified scene and part-parsing.
Also, \citet{de2021part} introduced a baseline model using a state-of-the-art panoptic segmentation network and a part segmentation network, merging them using heuristics.
The panoptic and part segmentation is merged in top-down or bottom-up manner.
In the top-down merge, the prediction from panoptic segmentation is re-used for scene-level semantic classes that do not consist of parts. Then for partitionable semantic classes, the corresponding segment of the part prediction is extracted.
In case of conflicting predictions, a void label will be assigned.
According to \citet{de2021part}, top-down merge produces better results than the bottom-up approach.
In addition, their paper has released two datasets with panoptic-part annotations: Cityscapes Panoptic Part (CPP) dataset and Pascal Panoptic Part (PPP) dataset \citep{meletis2020cityscapes}.
Along with the drawbacks of employing independent networks as mentioned in \cref{sec:intro}, there are concerns with the usage of top-down merge as shown in \cref{fig:teaser,fig:visual}.
Due to inconsistencies, top-down merging may result in undefined regions around the contours of objects. Due to some imbalance between \stuff\ and \things{}, it also has trouble separating them.
Our work resolves these issues by proposing a unified fusion for semantics, instances, and parts, giving equal priority to all individual predictions.

\begin{figure*}[t]
    \centering
    \includegraphics[width=0.9\linewidth]{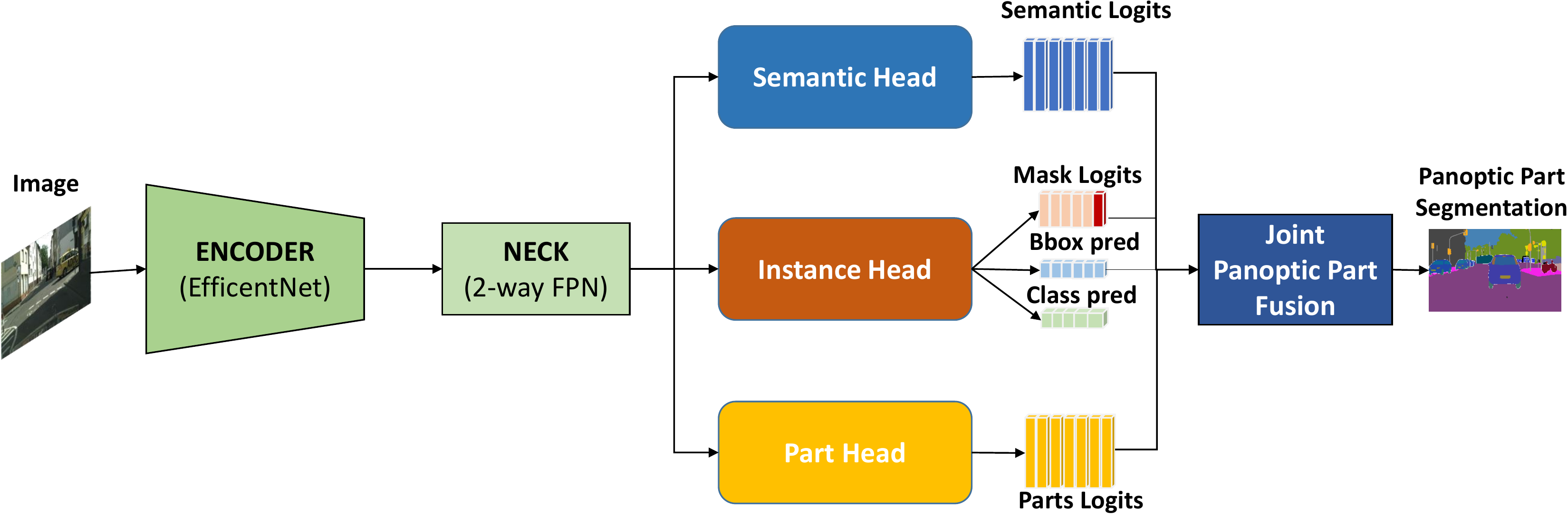}
    \caption{Overview of our proposed network architecture. It features a shared encoder, three specialized prediction heads, and a unified joint fusion module.}
    \label{fig:network}
\end{figure*}

\section{Unified \PPS{}} \label{sec:method}
Our work extends EfficientPS \citep{mohan2021efficientps} in two fundamental aspects: 1.) The network is  extended to incorporate a part segmentation head, and 2.) we propose our joint panoptic part fusion.

\subsection{Network Architecture} \label{sec:method:network}
We employ the backbone, semantic head, and instance head of EfficientPS \citep{mohan2021efficientps} in this work.
As part segmentation is regarded as a semantic segmentation problem, we are replicating the semantic branch of EfficientPS and train it for part-level segmentation.
All three resulting heads share a common EfficientNet-b5 backbone \citep{tan2019efficientnet}, which helps to ensure that the predictions made by the heads are consistent with one another.
The positive impact of the shared encoder is presented in \cref{sec:results:ablation}.
In order to produce \pps{}, we combine the predictions from all three heads in our proposed joint fusion.
The goal of \pps\ is to predict $(s,p,id)_i$ for each pixel $i$. Here, $s$ represents semantic scene level class from the semantic head, $p$ represents the part-level class and $id$ indicates the instance identifier which is obtained from the instance head. An overview of the architecture of our proposed model is shown in \cref{fig:network}.

\begin{figure*}[t]
    \centering
    \includegraphics[width=\linewidth]{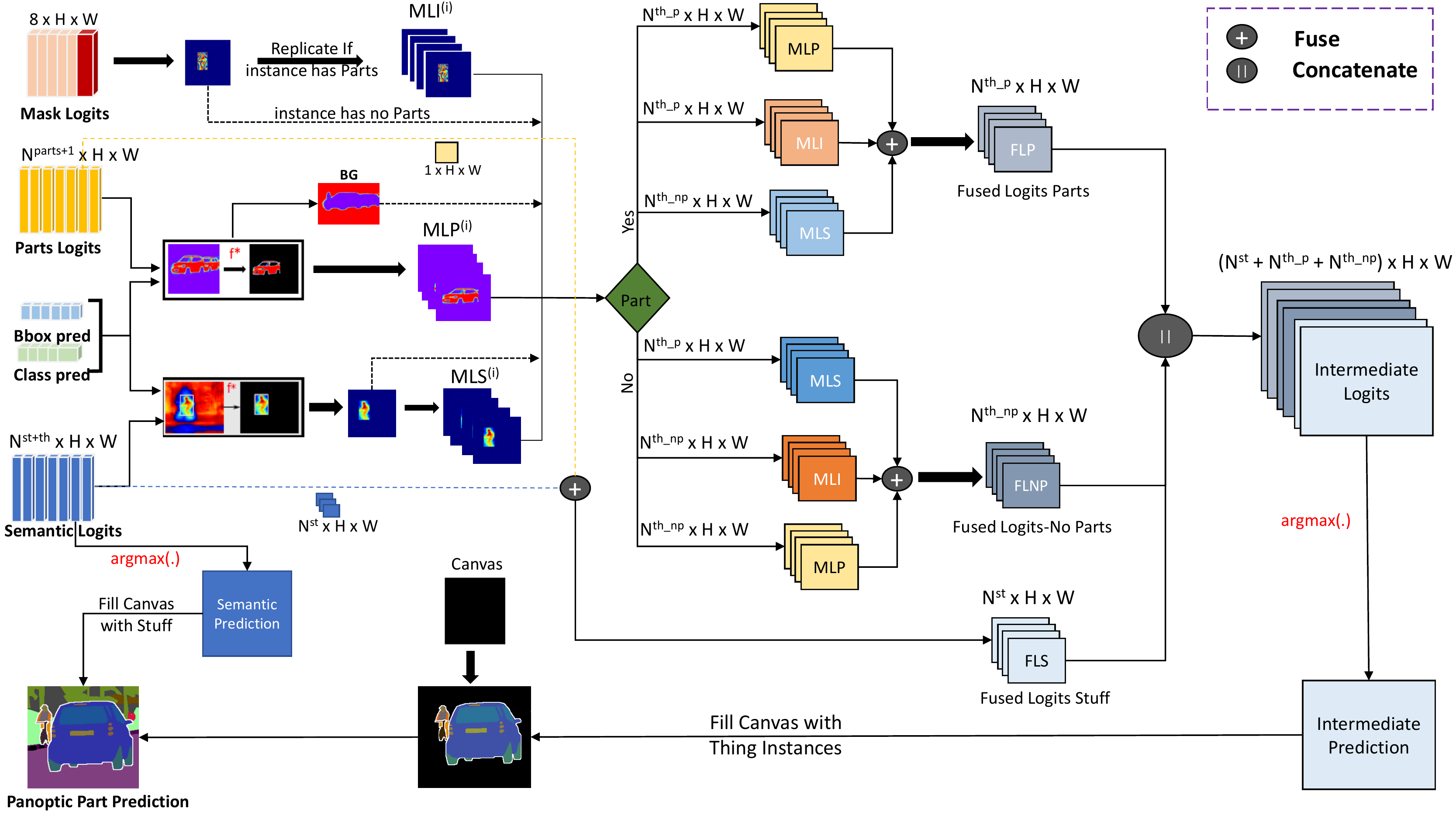}
     \caption{Illustration of our proposed joint fusion module. Semantic, instance, and part predictions are equally balanced and combined.}
     \label{fig:Fusion}
\end{figure*}

\subsubsection{Part Segmentation Head}  \label{sec:method:network:parts}
According to previous work \citep{de2021part}, the grouping of parts yields better results. \Ie, semantically identical parts, \eg the windows of cars or busses, are grouped into a single part class.
We have verified this finding for our network architecture (see \cref{tab:ablation2}) and consequently follow the same principle.
Another relevant design question for the part prediction head is concerned with the non-partitionable classes.
In our approach, we chose to represent all these classes as a single background class.
This avoids redundant predictions and further balances the learning of parts versus other classes.
Our decision is again validated by experiments of which the results are provided in \cref{tab:ablation2}.
Both groupings of classes (semantic grouping of parts, as well as grouping of the background) can later be easily distinguished by the additional information of the other prediction heads to obtain a fine-grained \pps.

\subsection{Joint Fusion} \label{sec:method:fusion}
To obtain \pps{}, one must combine the predictions of semantic segmentation, instance segmentation, and part segmentation.
In general, this includes four possible categories for fusion: Partitionable and non-partitionable \stuff{}, and partitionable and non-partitionable \things{}.
For the sake of verbosity, we only describe the three combinations which actually occur in the data (partitionable \stuff{} is not included), but our approach generalized to the last case as well.
Inspired by the panoptic fusion module of EfficientPS \citep{mohan2021efficientps}, we propose a joint panoptic part fusion module that fuses the individual results of the three heads by giving each prediction equal priority and thoroughly exploiting coherent predictions.
\cref{fig:Fusion} depicts our proposed joint panoptic part fusion module.

\paragraph{Fusion for \emph{Things}.}
The instance segmentation head predicts a set of object instances, each with its class prediction, confidence score, mask logits, and bounding box prediction.
The predicted instances are pre-filtered according to the steps carried out by EfficientPS \citep{mohan2021efficientps}, including confidence thresholding, non-maximum suppression, \etc.
After this, we obtain a bounding box, class prediction, and masked instance logits $MLI$ for every instance.
Simultaneously, we obtain the semantic logits of $N$ channels from the semantic head, where $N$ is the number of semantic classes, which is $N_{stuff}+N_{things}$.
Lastly, we obtain the part logits with $N_{P}$ channels from the part head, where $N_{P}$ is the number of grouped parts plus one additional channel for the background.
To balance the individual predictions, we normalize the semantic and part logits by applying a softmax function along the channel dimension.
In a next step, the appropriate channels of the semantic prediction is selected, based on the class prediction of each instance. This selected logits are further masked according to the instance's bounding box to yield the masked, semantic logits $MLS$.

Suppose the predicted class (by the instance head) is partitionable, then a subset of corresponding logits are selected from the part segmentation, \eg if the instance head predicts a person, the logits for head, torso, legs, and arms are selected. These logits are again masked by the corresponding bounding box to produce the third masked logits for parts $MLP$.
If the predicted class is not further segmentable into parts, the background class from the part logits is selected instead and masked likewise.
To make the fusion operation feasible, we replicate $MLS$ and $MLI$ to match the number of corresponding parts.
For example, a person instance contains four parts (head, arms, torso, legs), thus $MLP$ is of shape  $4 \times W \times H$. Therefore, $MLS$ and $MLI$ are replicated 4 times to match the shape of $MLP$.

By now, three sets of masked logits are available.
We are now fusing these logits separately for classes with and without parts in the same fashion.
To compute the fused logits for classes with parts $FLP$ and class without parts $FLNP$, we form the sum of the sigmoid of the masked logits and the sum of the masked logits and compute the Hadamard product of both.
This procedure is depicted in \cref{eq:fusion}:
\begin{equation}
FL\left(MLL\right)=\left(\sum_{l \in MLL}\sigma(l)\right) \odot \left(\sum_{l \in MLL} l\right)
\label{eq:fusion}
\end{equation}
In this equation, $\sigma(\cdot)$ denotes the sigmoid function, $\odot$ denotes the Hadamard product, and $MLL$ is a set of masked logits which are supposed to be fused, \eg $MLL = \{MLS, MLI, MLP\}$.
This equation describes a generalized version of the fusion proposed by \citet{mohan2021efficientps} that handles arbitrarily many logits.

\paragraph{Fusion for \emph{Stuff}.}
To generate the fused logits $FLS$ for the \stuff\ classes, the $N_{stuff}$ channels from the semantic head are fused with the background channel of the part head in the same manner, \ie according to \cref{eq:fusion}, but this time with only two sets of logits (no instance information).
As mentioned, the same concept would also apply for \stuff{} that is partitionable.

\paragraph{Overall Fusion.}
All three fused logits, $FLP$, $FLNP$, and $FLS$, are concatenated along the channel dimensions to obtain intermediate logits, which produce the intermediate panoptic-part prediction by taking the \textit{argmax} of these intermediate logits.
Finally, we fill an empty canvas with the intermediate panoptic-part prediction for all \things.
The remaining empty parts, \ie the background, of the canvas is filled with the prediction for \stuff{} classes extracted from the semantic segmentation head.
Lastly, \stuff{} areas below a minimum threshold $min_{stuff}$ are filtered, as by \citet{mohan2021efficientps}.
During fusion, the fused score increases if the predictions of all three heads are consistent, and likewise it is decreased if the predictions do not match with eachother.

\section{Experiments and Results} \label{sec:results}
\paragraph{Datsets.}
As mentioned before, we use the recently introduced Cityscapes Panoptic Parts (CPP) and Pascal Panoptic Parts (PPP) datasets \citep{meletis2020cityscapes}.
CPP provides pixel-level annotations for 11 \stuff\ classes and 8 \things\ classes, totaling 19 object classes.
Out of the 8 \things{}, five include annotations at the part level.
There are 2975 images for training and 500 for validation in this finely annotated dataset.
PPP consists of 100 object classes, with 20 \things\ and 80 \stuff\ classes.
Part-level annotations are present in 16 of the 20 \things. As in previous work \citep{meletis2020cityscapes}, we only consider a subset of 59 object classes for training and evaluation, including 20 \things\ and 39 \stuff\ classes, and 58 part classes.
These parts are detailed by \citet{michieli2020gmnet} and \citet{zhao2019multi}.
PPP consists of a total of 10103 images which are divided into 4998 images for training and 5105 for validation.

\paragraph{Training Details.}
For the Cityscapes data, we use images of the original resolution, \ie $1024\times2048$ pixels, and resize the input images of PPP to $384\times512$ pixels for training.
We perform data augmentation, scaling and hyperparameter initialization as in EfficientPS \citep{mohan2021efficientps}.
We use a multi-step learning rate ($lr$) and train our network by Stochastic Gradient Descent (SGD) with a momentum of $0.9$.
For the CPP and PPP, we use a start $lr$ of 0.07 and 0.01, respectively.
We begin the training with a warm-up phase in which the $lr$ is increased linearly from $\frac{1}{3}lr$ up to $lr$ within $200$ iterations.
The weights of all InPlace-ABN layers \citep{bulo2018place} are frozen, and we train the model for $10$ additional epochs with a fixed learning rate of $10^{-4}$. Finally, we unfreeze the weights of the InPlace-ABN layers and train the model for $50k$ iterations beginning with $lr$ of 0.07 (CPP) and 0.01 (PPP), and reduce $lr$ after iterations $32k$ and $44k$ by a factor of 10.
Four GPUs are used for the training with a batch size of $2$ per GPU for CPP and $8$ per GPU for PPP.

\paragraph{Metrics.}
In this paper, we evaluate the individual semantic and part segmentation using mean Intersection over Union (mIoU), and the instance segmentation using mean Average Precision (mAP).
For the evaluation of our \pps{}, we use the Part Panoptic Quality (PartPQ) \citep{de2021part}, which is an extension of the Panoptic Quality (PQ) \citep{kirillov2019panoptic}.

\subsection{Comparison to State-of-the-Art} \label{sec:results:sota}
  
The baseline approach by \citet{de2021part} uses the panoptic labels of the Cityscapes dataset \citep{cordts2016cityscapes} to train a panoptic segmentation network.
Since this data is slightly different from the recently annotated panoptic part dataset (CPP) presented by \citet{de2021part}, a direct, fair comparison is not possible. \cref{tab:baselinecomparision} clearly demonstrates that the CPP dataset differs, as the introduction of parts has resulted in inconsistencies of annotations.
To make the baseline comparable to our approach in terms of data, we re-implement the baseline and train it on the same data.
The re-implementation consists of EfficientPS \citep{mohan2021efficientps} for panoptic segmentation, and our part segmentation network with a separate backbone (\cf \cref{sec:method:network:parts}).
Top-down merging is then used to combine the two independent results into a \pps.
Our model is compared to the reproduced baseline and the official baseline of \citet{de2021part}. The official baseline consists of EfficientPS \citep{mohan2021efficientps} and BSANet \citep{zhao2019multi} with top-down merging.
The results of this comparison are shown in \cref{tab:sota} for single-scale and multi-scale inference.

\begin{table}[t]
    \centering
    \caption{Comparison of EfficientPS \citep{mohan2021efficientps} trained on cityscapes panoptic dataset with EfficientPS trained with Cityscapes Panoptic Part (CPP) dataset \citep{de2021part} and single-scale testing. $^\ast$ indicates the model trained with CPP dataset.}
    \label{tab:baselinecomparision}
    \begin{tabular}{c|ccc}
       \begin{tabular}{c}
           Method \\
       \end{tabular}
         & \begin{tabular}{c} PQ \end{tabular} & \begin{tabular}{c} SQ \end{tabular} & \begin{tabular}{c} RQ \end{tabular}\Bstrut\\
        \hline
        EfficientPS & 63.9 & 81.5 & 77.1\Tstrut\\
        EfficientPS$^\ast$ & 62.2 & 81.0 & 75.7
        
    \end{tabular}
\end{table}  

\begin{table*}[t]
    \centering
    \caption{Evaluation results of \pps{} on Cityscapes and Pascal Panoptic Parts \citep{meletis2020cityscapes} compared to state-of-the art. \textit{P} and \textit{NP} refer to areas with and without part labels, respectively. $^\ast$ indicates our reproduced baseline (details in \cref{sec:results:sota}). $\dagger$ indicates that the number of parameters refer to the encoders only.}
    \label{tab:sota}
    \begin{tabular}{c|ccc|ccc|ccc}
        \multirow{3}{*}{Method} & \multicolumn{3}{c|}{Before Merge/Fusion} & \multicolumn{3}{c|}{After Merge/Fusion} & \multirow{3}{*}{\begin{tabular}{c} Density\\{[\%]} \end{tabular}} & \multirow{3}{*}{\begin{tabular}{c} Run\\time\\{[ms]} \end{tabular}} & \multirow{3}{*}{\begin{tabular}{c} Model\\size\\{[M]} \end{tabular}}\\
         & \multirow{2}{*}{\begin{tabular}{c} Sem.\\mIoU \end{tabular}} & \multirow{2}{*}{\begin{tabular}{c} Inst.\\AP \end{tabular}} & \multirow{2}{*}{\begin{tabular}{c} Part\\mIoU \end{tabular}} & \multicolumn{3}{c|}{PartPQ} & & & \\
         & & & & All & P & NP & \Bstrut\\
        \hline
        \multicolumn{10}{c}{Cityscapes Panoptic Parts, Single-Scale}\Tstrut\Bstrut\\
        \hline
        Baseline$^\ast$ & 79.7 & 36.6 & 74.5 & 57.7 & 44.2 & 62.5 & 98.84 & 871 & 68.8\Tstrut\\
        \ours & 80.5 & 37.9 & 77.0 & 59.6 & 47.7 & 63.8 & 99.33 & 397 & 44.19 \Bstrut\\
        \hline
        \multicolumn{10}{c}{Cityscapes Panoptic Parts, Multi-Scale}\Tstrut\Bstrut\\
        \hline
        Baseline & 80.3 & 39.7 & 76.0 & 60.2 & 46.1 & 65.2 & -- & -- & --\Tstrut\\
        \ours  & 81.8 & 41.3 & 78.5 & 61.8 & 50.8 & 65.7 & 99.50 & 2498 & 44.19 \Bstrut\\
        \hline
        \multicolumn{10}{c}{Pascal Panoptic Parts, Single-Scale}\Tstrut\Bstrut\\
        \hline
      Baseline-1
       & 47.1 & 38.5 & 53.9 & 31.4 & 47.2 & 26.0 & -- & -- & 68$^{\dagger}$\Tstrut\\
        Baseline-2 & 55.1 & 44.8 & 58.6 & 38.3 & 51.6 & 33.8 & -- & -- & 111$^{\dagger}$\\
        \ours & 46.0 & 39.1 & 54.4 & 32.3 & 48.3 & 26.9 & 92.10 & 146 & 44.19
    \end{tabular}
\end{table*}

For CPP, the results indicate that our proposed network improves accuracy significantly compared to the reproduced baseline for single-scale testing.
Our JPPF outperforms the reproduced baseline by 1.9 percentage points (pp) in overall $Part PQ$ and significantly by 3.5 pp in ${Part PQ}_P$.
Similarly for multi-scale testing, our proposed model outperforms the baseline by 1.6 pp and 4.7 pp in $Part PQ$ and ${Part PQ}_P$, respectively.
Furthermore, our model betters both baselines in all individual predictions before merging/fusion.
In addition, JPPF produces denser results than the baseline, which enhances the density by 0.5 pp for single-scale testing and by 0.66 pp for multi-scale testing.

For PPP, our model outperforms the top-down combination DeepLabV3+ \citep{chen2018b} and Mask RCNN \citep{HeGDG17} (\textit{Baseline-1}), even though this baseline was trained with the original Pascal parts and Pascal panoptic segmentation datasets, which provide more annotations.
\textit{Baseline-2} (top-down merging of DLv3-ResNeSt269 \citep{ChenPSA17, zhang2022resnest}, DetectoRS \citep{qiao2020detectors}, and BSANet \citep{zhao2019multi}) obtains an even better result because it is constructed from much more complex models, and hence has a higher representational capacity.
However, when comparing the model size (see \cref{tab:sota}), it shows that the backbone of \textit{Baseline-2} alone is already more than two times larger than our whole model.
%

\begin{figure*}[t]
    \centering
    \begin{subfigure}{0.245\linewidth}
\includegraphics[width=\linewidth]{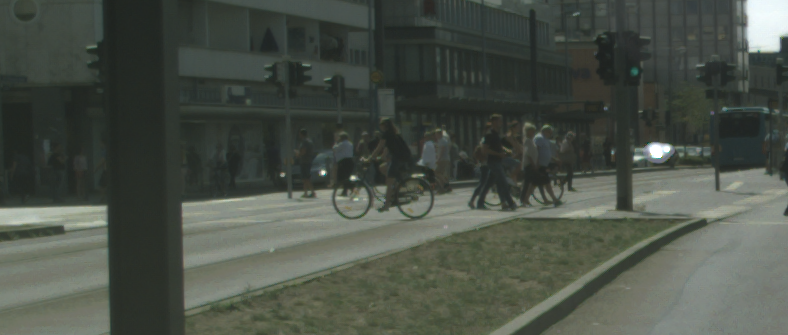}
    \end{subfigure}\hspace*{\fill}
     \begin{subfigure}{0.245\linewidth}
\includegraphics[width=\linewidth]{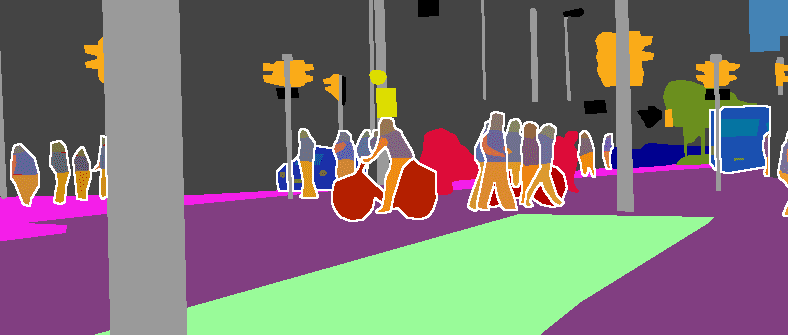}
    \end{subfigure}\hspace*{\fill}
    \begin{subfigure}{0.245\linewidth}
\includegraphics[width=\linewidth]{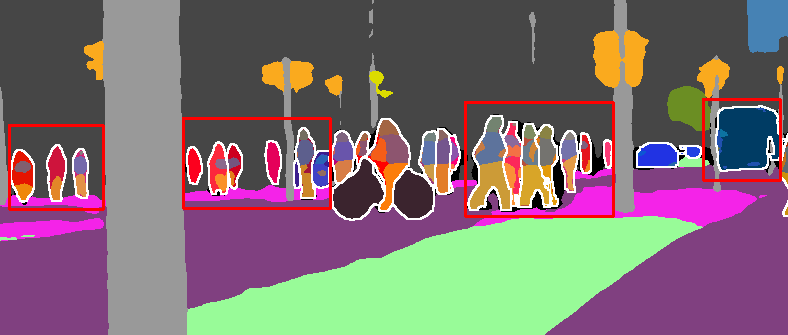}
    \end{subfigure}\hspace*{\fill}
     \begin{subfigure}{0.245\linewidth}
\includegraphics[width=\linewidth]{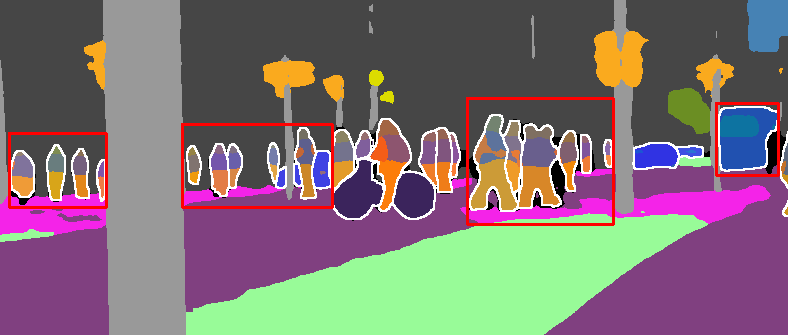}
    \end{subfigure}
    
    \vspace*{\fill}
    
    \begin{subfigure}{0.245\linewidth}
\includegraphics[width=\linewidth]{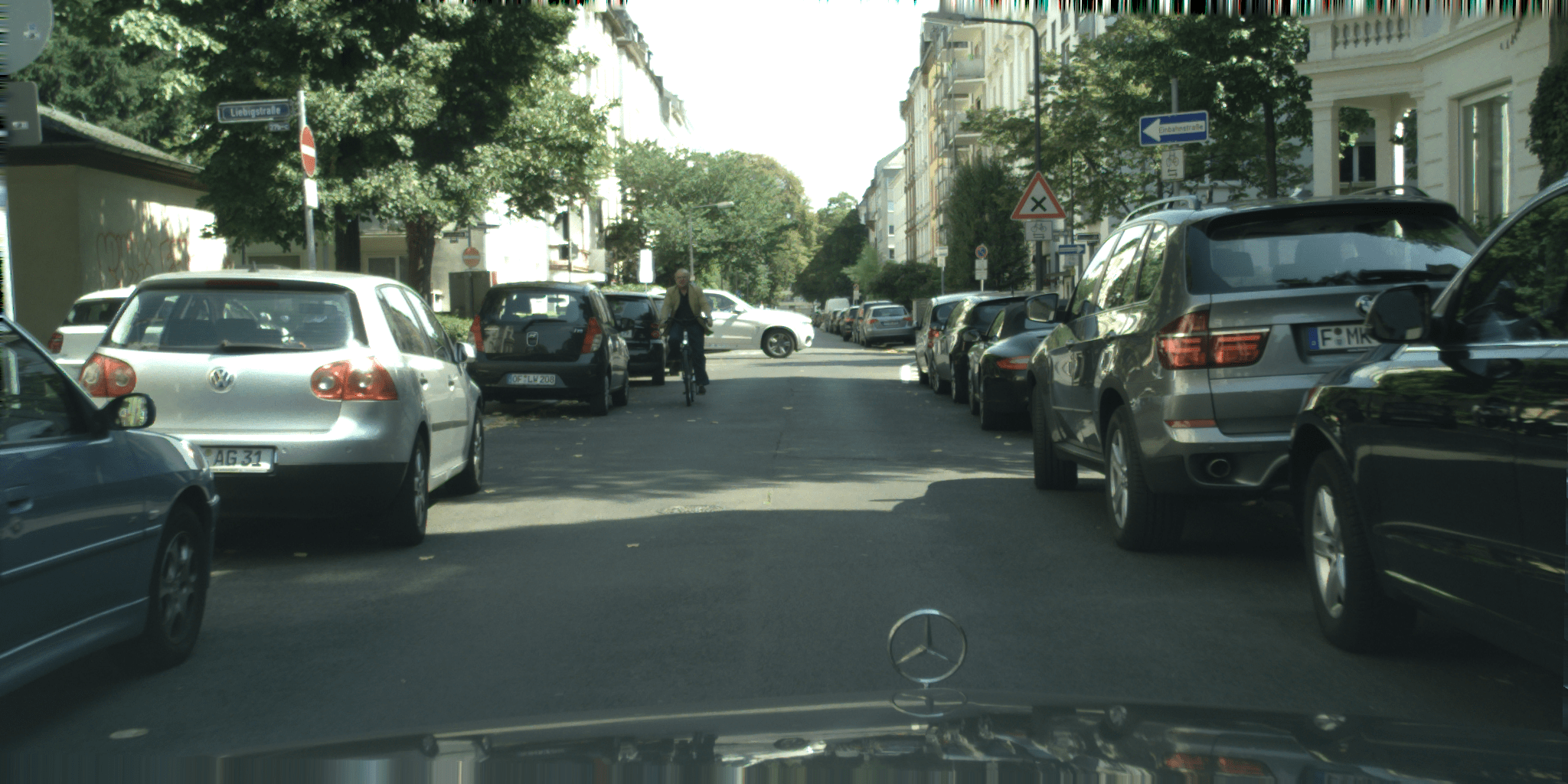}
    \end{subfigure}\hspace*{\fill}
     \begin{subfigure}{0.245\linewidth}
\includegraphics[width=\linewidth]{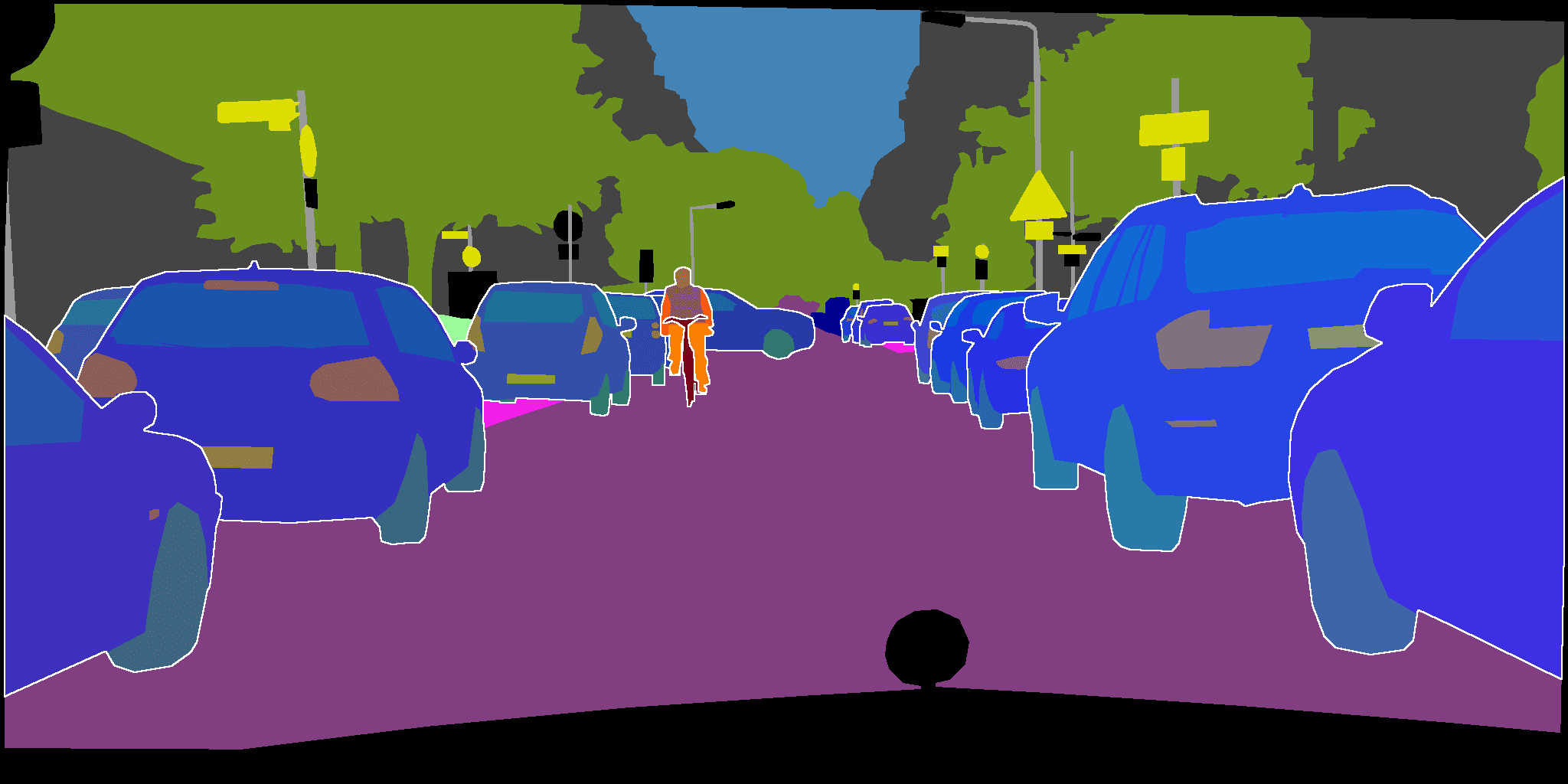}
    \end{subfigure}\hspace*{\fill}
    \begin{subfigure}{0.245\linewidth}
\includegraphics[width=\linewidth]{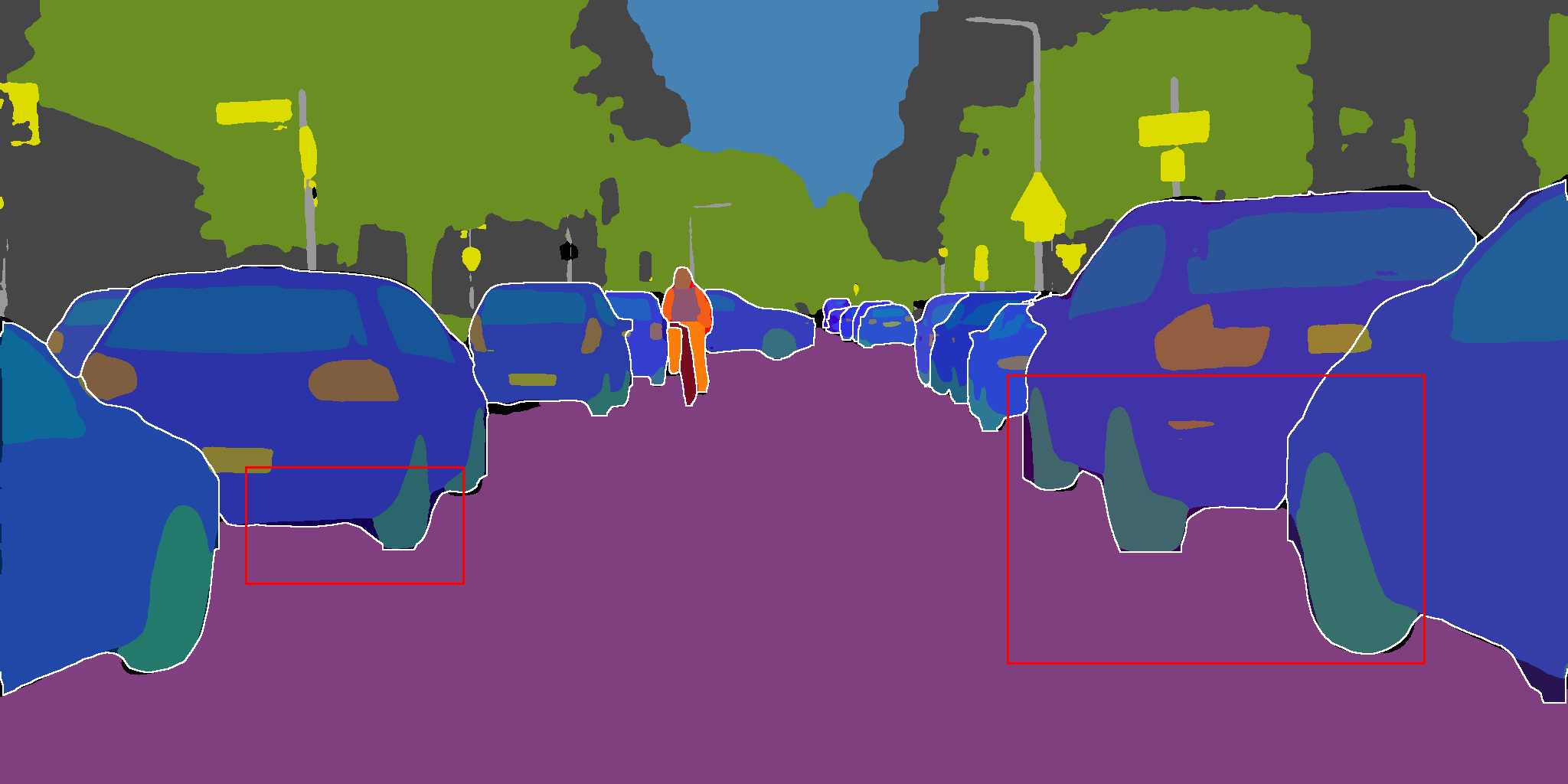}
    \end{subfigure}\hspace*{\fill}
     \begin{subfigure}{0.245\linewidth}
\includegraphics[width=\linewidth]{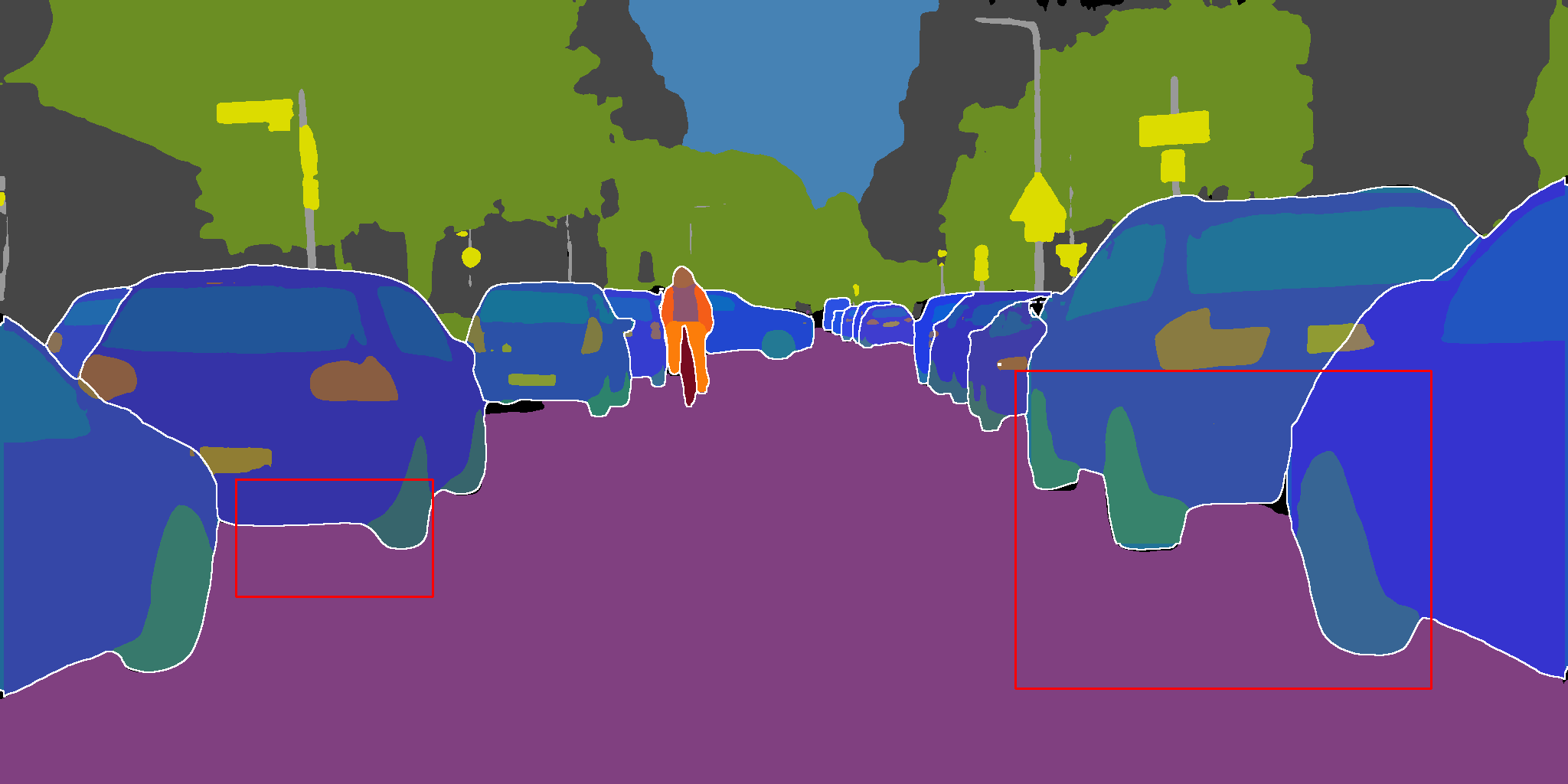}
    \end{subfigure}
    
    \vspace*{\fill}
    \begin{subfigure}{0.245\linewidth}
\includegraphics[width=\linewidth]{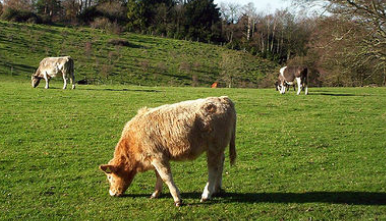}

    \end{subfigure}\hspace*{\fill}
     \begin{subfigure}{0.245\linewidth}
\includegraphics[width=\linewidth]{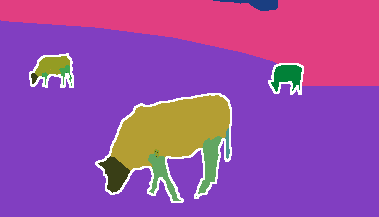}

    \end{subfigure}\hspace*{\fill}
    \begin{subfigure}{0.245\linewidth}
\includegraphics[width=\linewidth]{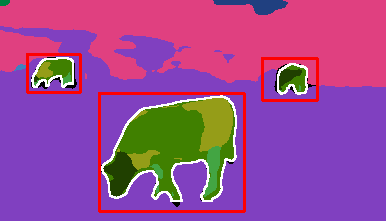}

    \end{subfigure}\hspace*{\fill}
     \begin{subfigure}{0.245\linewidth}
\includegraphics[width=\linewidth]{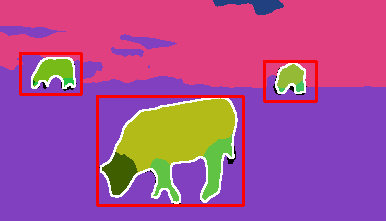}

    \end{subfigure}

    \vspace*{\fill}
   \begin{subfigure}{0.245\linewidth}
\includegraphics[width=\linewidth]{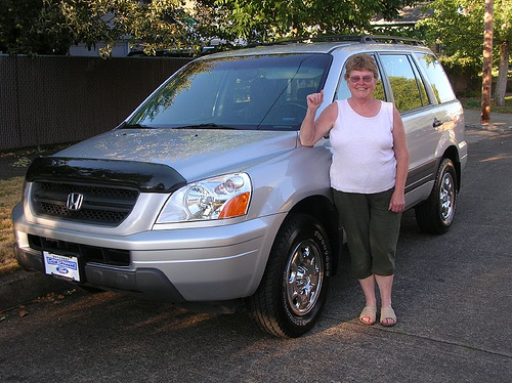}
\subcaption*{Original Image}
    \end{subfigure}\hspace*{\fill}
     \begin{subfigure}{0.245\linewidth}
\includegraphics[width=\linewidth]{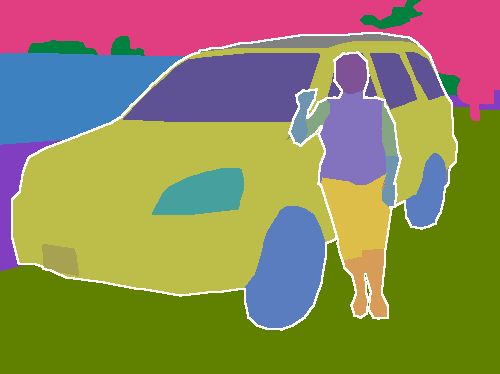}
\subcaption*{Ground-truth}
    \end{subfigure}\hspace*{\fill}
    \begin{subfigure}{0.245\linewidth}
\includegraphics[width=\linewidth]{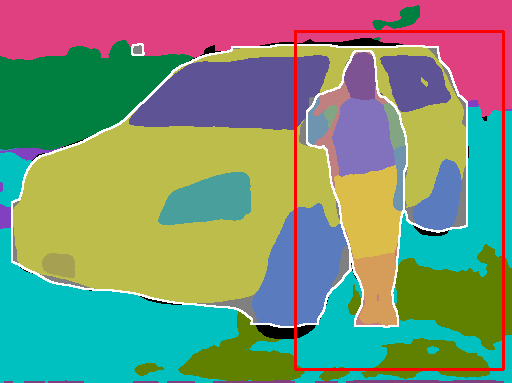}
\subcaption*{Baseline$^\ast$}
    \end{subfigure}\hspace*{\fill}
     \begin{subfigure}{0.245\linewidth}
\includegraphics[width=\linewidth]{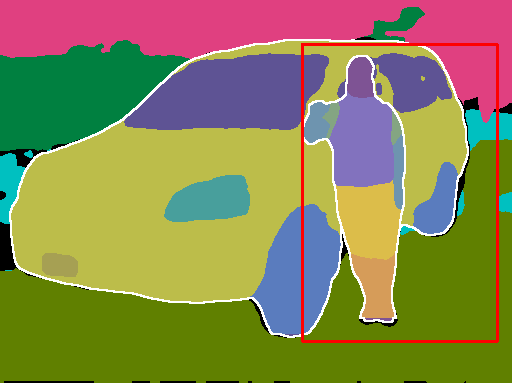}
\subcaption*{\ours}
    \end{subfigure}
    
    \caption{Qualitative results of our proposed model on Citscapes and Pascal Panoptic Parts compared to our reproduced baseline, ground-truth and the reference image.
    More visual examples for both datasets are provided in the appendix in \cref{fig:visualcityscapes,fig:visualpascal}.}
    \label{fig:visual}
\end{figure*}

From \cref{fig:visual}, we can see that our proposed fusion is able to segment the parts of very small and distant object classes reliably.
Also, our proposed fusion solves the typical problems of top-down merging (\cf \cref{sec:intro}).
As illustrated in \cref{fig:visual}, there are no unknown regions within objects (\things), since our fusion gives equal priority to all three heads.
The second issue of \stuff{} classes bifurcating \things{} (as shown in \cref{fig:teaser}) is also improved largely.
This is due to the introduction of fusion between \stuff{} classes of semantic logits and the background class of part logits.
Lastly when comparing the model sizes and inference times, we can highlight another advantage of our unified model: It is more efficient as it requires fewer parameters. 
On average, the inference per image requires only 397ms, which is less than half of the time required by the baseline.

\begin{table}[t]
    \centering
    \caption{Ablation Study on Cityscapes Panoptic Parts. The design choices of our part segmentation head are validated, and we contrast independent and shared feature encoders.}
    \label{tab:ablation2}
    \resizebox{\linewidth}{!}{
    \begin{tabular}{c|ccc}
       \begin{tabular}{c}
           Method \\
       \end{tabular}
         & \begin{tabular}{c} Sem.\\mIoU \end{tabular} & \begin{tabular}{c} Inst.\\AP \end{tabular} & \begin{tabular}{c} Part\\mIoU \end{tabular}\Bstrut\\
        \hline
        Grouped Parts & -- & -- & 74.5\Tstrut\\
        Non-Grouped Parts & -- & -- & 65.7\Bstrut\\
        \hline 
        Grouped Parts + SemBG & -- & -- & 75.6\Tstrut\\
        Grouped Parts + BG (Ours)  & -- & -- & 77.0\Bstrut\\
        \hline
        Independent Networks & 78.1 & 37.3 & 74.5\Tstrut\\
        Shared Features (Ours) & 80.5 & 37.9 & 77.0\\

    \end{tabular}
}
\end{table}

\begin{table*}[hbt!]
    \centering
    \caption{Ablation Study on Cityscapes and Pascal Panoptic Parts \citep{meletis2020cityscapes}. We compare the uni-directional top-down merge to our proposed joint fusion module.}
    \label{tab:ablation1}
    \begin{tabular}{c|ccc|ccc|c}
        \multirow{3}{*}{Method} & \multicolumn{3}{c|}{Before Merge/Fusion} & \multicolumn{3}{c|}{After Merge/Fusion} & \multirow{3}{*}{\begin{tabular}{c} Density\\{[\%]} \end{tabular}}\\
         & \multirow{2}{*}{\begin{tabular}{c} Sem.\\mIoU \end{tabular}} & \multirow{2}{*}{\begin{tabular}{c} Inst.\\AP \end{tabular}} & \multirow{2}{*}{\begin{tabular}{c} Part\\mIoU \end{tabular}} & \multicolumn{3}{c|}{PartPQ} & \\
         & & & & All & P & NP &\Bstrut\\
        \hline
        \multicolumn{8}{c}{Cityscapes Panoptic Parts, Single-Scale}\Tstrut\Bstrut\\
        \hline
        Ours w/ Top-Down-Merge & 80.5 & 37.9 & 77.0 & 59.5 & 47.5 & 63.7 & 99.13 \\ 
        \ours & 80.5 & 37.9 & 77.0 & 59.6 & 47.7 & 63.8 & 99.33\Tstrut\\
        \hline
        \multicolumn{8}{c}{Cityscapes Panoptic Parts, Multi-Scale}\Tstrut\Bstrut\\
        \hline
        Ours w/ Top-Down-Merge & 81.8 & 41.3 & 78.5 & 61.6 & 50.7 & 65.5 & 99.20 \\
        \ours & 81.8 & 41.3 & 78.5 & 61.8 & 50.8 & 65.7 & 99.50\Tstrut\\
        \hline
        \multicolumn{8}{c}{Pascal Panoptic Parts, Single-Scale}\Tstrut\Bstrut\\
        \hline
        Ours w/ Top-Down-Merge & 46.0 & 39.1 & 54.4 & 29.0 & 37.8 & 26.0 & 89.57\Tstrut\\
        \ours & 46.0 & 39.1 & 54.4 & 32.3 & 48.3 & 26.9 & 92.10  
        
    \end{tabular}
\end{table*}

\begin{table*}[hbt!]
    \centering
    \caption{Run-time comparison of JPPF to the baseline on Cityscapes Panoptic Parts. $^\ast$~indicates the reproduced baseline which is detailed in \cref{sec:results:sota}.}
    \label{tab:ablation3}
    \begin{tabular}{c|c|ccc|c}
        \multirow{3}{*}{Method} & \multirow{3}{*}{\begin{tabular}[c]{@{}c@{}}Individual\\Predictions\\ {[ms]}\end{tabular}} & \multicolumn{3}{c|}{Fuse/Merge {[ms]}}                                & \multirow{3}{*}{\begin{tabular}[c]{@{}c@{}}Total\\Inference\\ {[ms]}\end{tabular}} \\ 
         &   & \begin{tabular}[c]{@{}c@{}}Panoptic\\Fusion\end{tabular} & Merge & \begin{tabular}[c]{@{}c@{}}Joint\\Fusion\end{tabular} &\Bstrut\\ 
        \hline
        Baseline$^\ast$  & 269 & 118  & 484   & --  & 871\Tstrut\\
        Ours w/ Merge  & 215 & 118  & 484   & --  & 817 \\
        \ours        & 236      & --  & -- & 161 & 397
\end{tabular}%
\end{table*}

\subsection{Ablation Study} \label{sec:results:ablation}
\subsubsection{Shared Encoder vs. Independent Encoders}
Our aim is to jointly learn semantic, instance, and part segmentation in a single, unified model.
To validate that these three tasks benefit from a common feature representation, we compare our results before fusion to three separate equivalent networks that have been trained individually with different encoders.
The model with a single, shared encoder surpasses the individual models in all three tasks (see \cref{tab:ablation2}).
The improvement is 2.4 pp, 2.5 pp, and 0.6 pp for semantic, part, and instance segmentation, respectively.
This result clearly indicates that using a shared encoder enables the network to learn a common feature representation, resulting in more accurate individual outcomes of each head.


\subsubsection{Top-down Merge vs. Joint Fusion}
Next, we compare our joint fusion module to the previously presented top-down merging strategy \citep{de2021part} in \cref{tab:ablation1}.
The proposed fusion module surpasses the top-down merge in terms of $PartPQ$, ${Part PQ}_{P}$, ${Part PQ}_{NP}$ in all test settings.
Even though our proposed fusion is admittedly only slightly better, the joint fusion produces also denser results than the uni-directional merge, indicating the improved consistency before and after fusion.
Additionally and as explained earlier, our fusion resolves the typical issues that are present with the top-down merge, as seen in \cref{fig:visual,fig:teaser}.
This is achieved by incorporating the part prediction into a mutual fusion, and mainly reflected for the results in areas that are partitionable.
Since the \things{} with part labels are limited in CPP, the impact is best observed on the PPP dataset.
On this data, our proposed fusion module is significantly better.
Specifically ${Part PQ}_{P}$ is improved by 10.5 pp, by giving equal priority to the parts during fusion.

\subsection{Run-time Analysis} \label{sec:results:runtimeanalysis}
We further assessed the efficiency of our proposed model with joint fusion, and the results are displayed in \cref{tab:ablation3}.
It is evident that the top-down merging requires more than twice the time compared to our proposed fusion.
To obtain \pps{} as proposed by \citet{de2021part}, one must first perform a panoptic fusion and then combine it with the part segmentation, which adds an extra overhead.
In comparison to the baseline, our approach is even more efficient because it uses a single backbone.

\section{Conclusion} \label{sec:conclusion}
In this paper, we proposed a unified network that helps to generate semantic, instance, and part segmentation and effectively combines them to provide a consistent \pps{}.
Our proposed model with joint fusion significantly outperforms the state-of-the-art by 1.6 pp in overall $PartPQ$ and by 4.7 pp in $PartPQ_P$ on the CPP dataset.
For the PPP dataset, our model with joint fusion outperforms our model with the top-down merge significantly by 3.3 pp in overall $PartPQ$ and by 10.5 pp in $PartPQ_P$.
With the addition of \stuff{} and parts into the fusion, our suggested fusion modules addresses the problems encountered in the top-down merge, such as unknown pixels inside contours and the bifurcation of \things\ and \stuff.
When compared to top-down merge, our suggested joint fusion is faster and produces denser results with superior segmentation quality.

For future work, we plan to interpolate the remaining, filtered regions in the prediction to obtain a fully dense \pps{}.

\section*{\uppercase{Acknowledgements}}
This work was partially funded by the Federal Ministry of Education and Research Germany under the project DECODE (01IW21001).

\bibliographystyle{apalike}
{\small
\bibliography{example}}

\section*{\uppercase{Appendix}}
\begin{figure*}[b!]
    \centering
    \begin{subfigure}{0.33\linewidth}
\includegraphics[width=\linewidth]{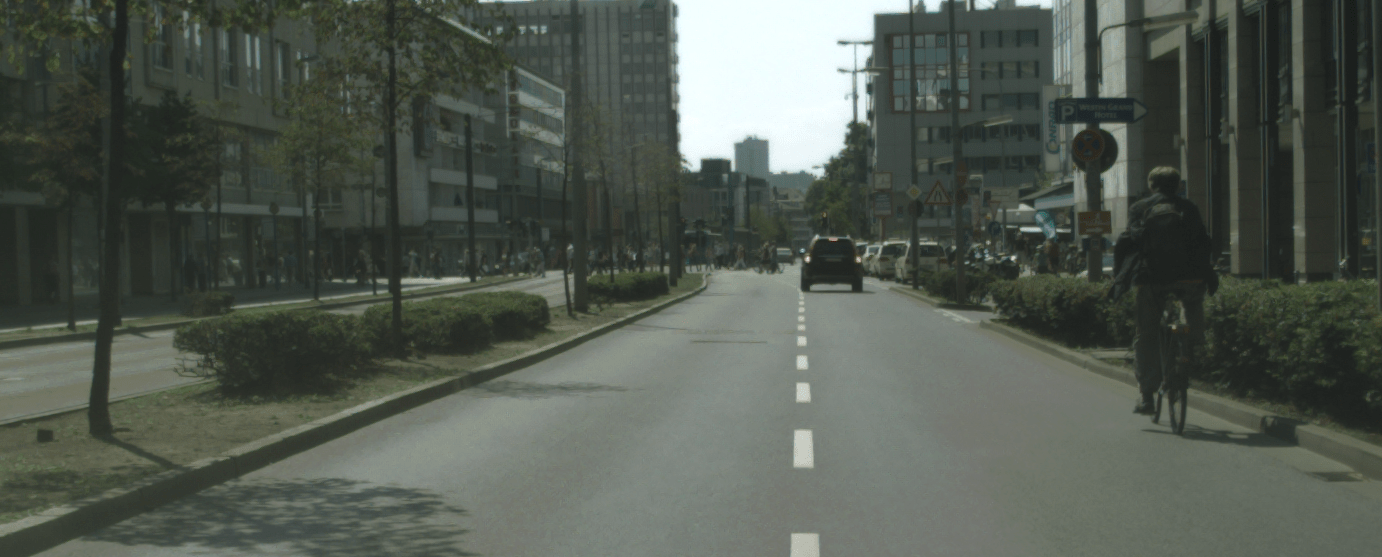}
    \end{subfigure}\hspace*{\fill}
     \begin{subfigure}{0.33\linewidth}
\includegraphics[width=\linewidth]{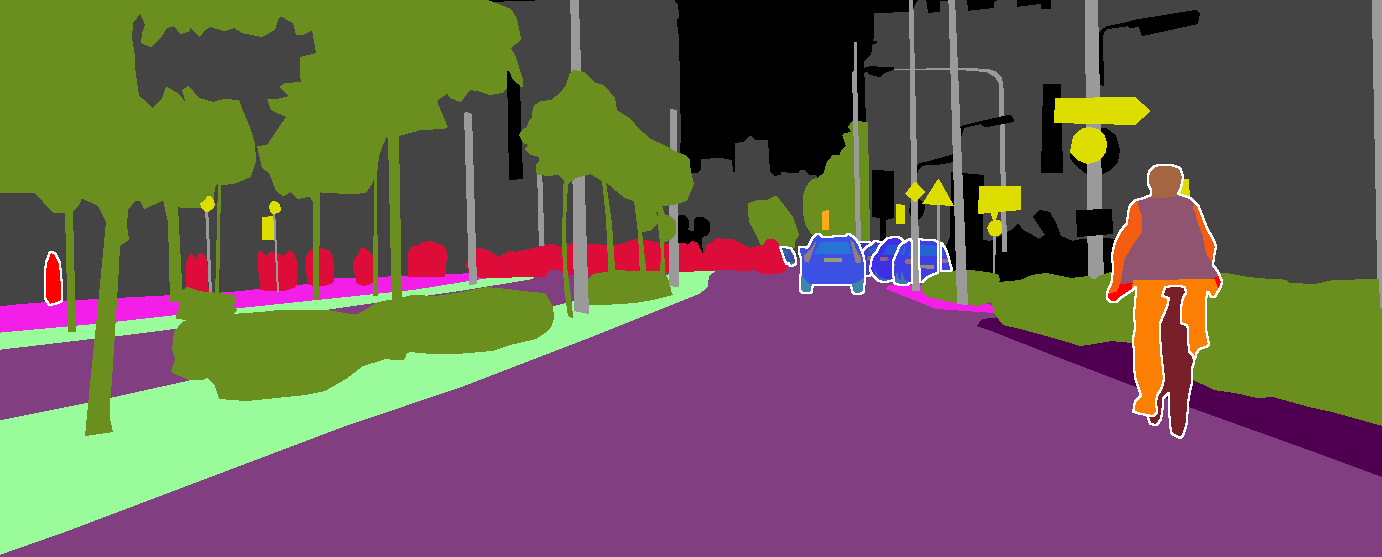}
    \end{subfigure}\hspace*{\fill}
     \begin{subfigure}{0.33\linewidth}
\includegraphics[width=\linewidth]{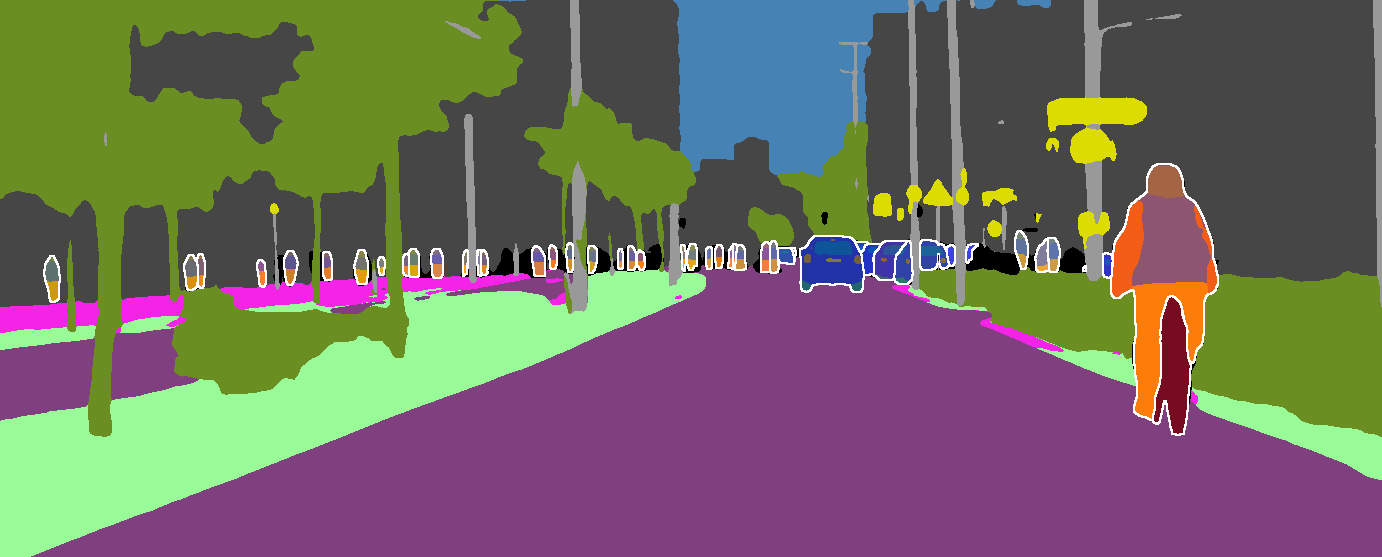}
    \end{subfigure}
    
    \vspace*{\fill}

            \begin{subfigure}{0.33\linewidth}
\includegraphics[width=\linewidth]{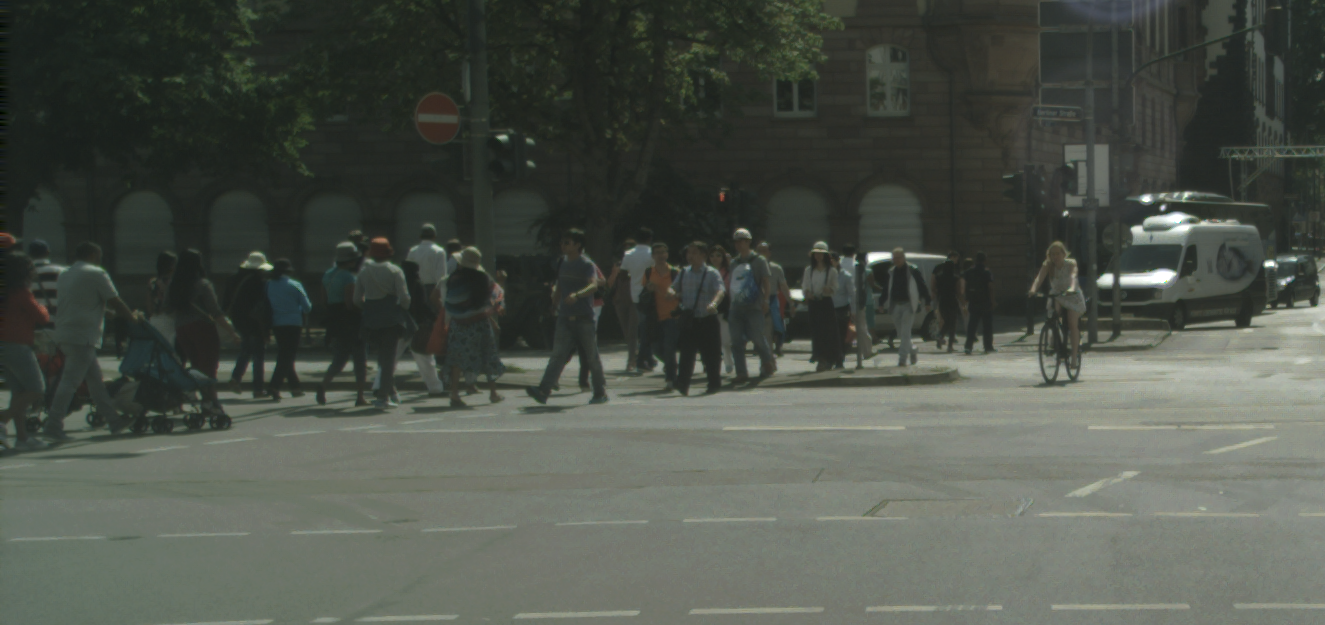}
    \end{subfigure}\hspace*{\fill}
     \begin{subfigure}{0.33\linewidth}
\includegraphics[width=\linewidth]{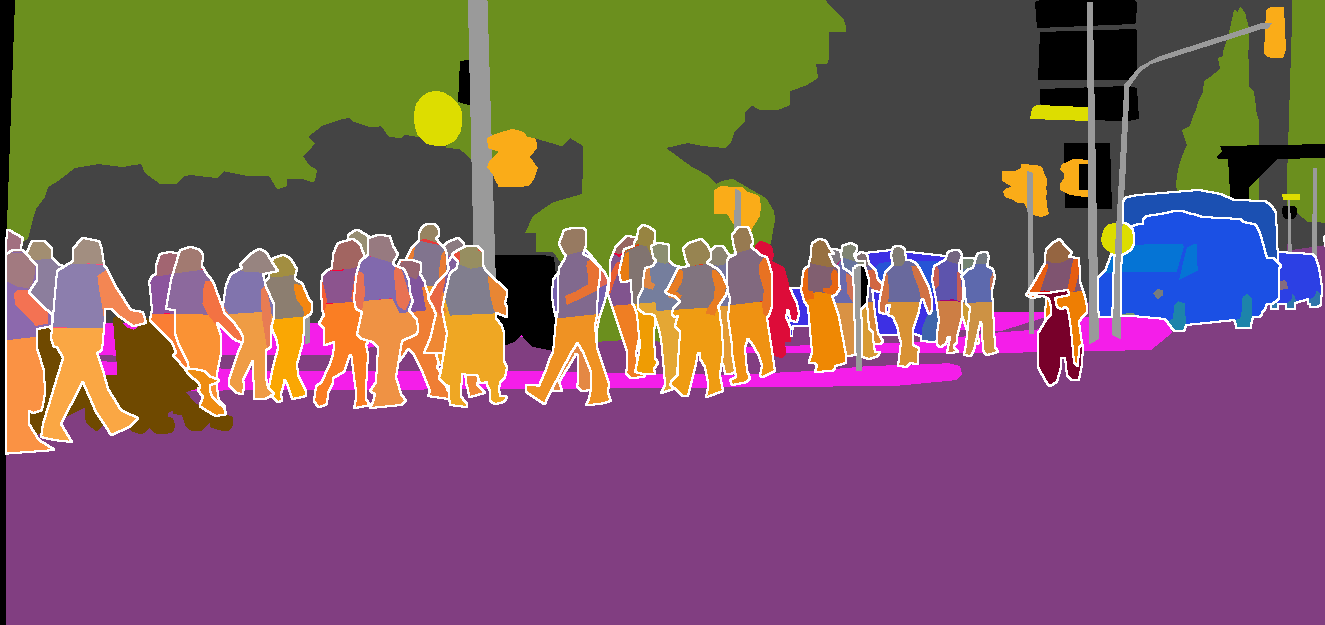}
    \end{subfigure}\hspace*{\fill}
     \begin{subfigure}{0.33\linewidth}
\includegraphics[width=\linewidth]{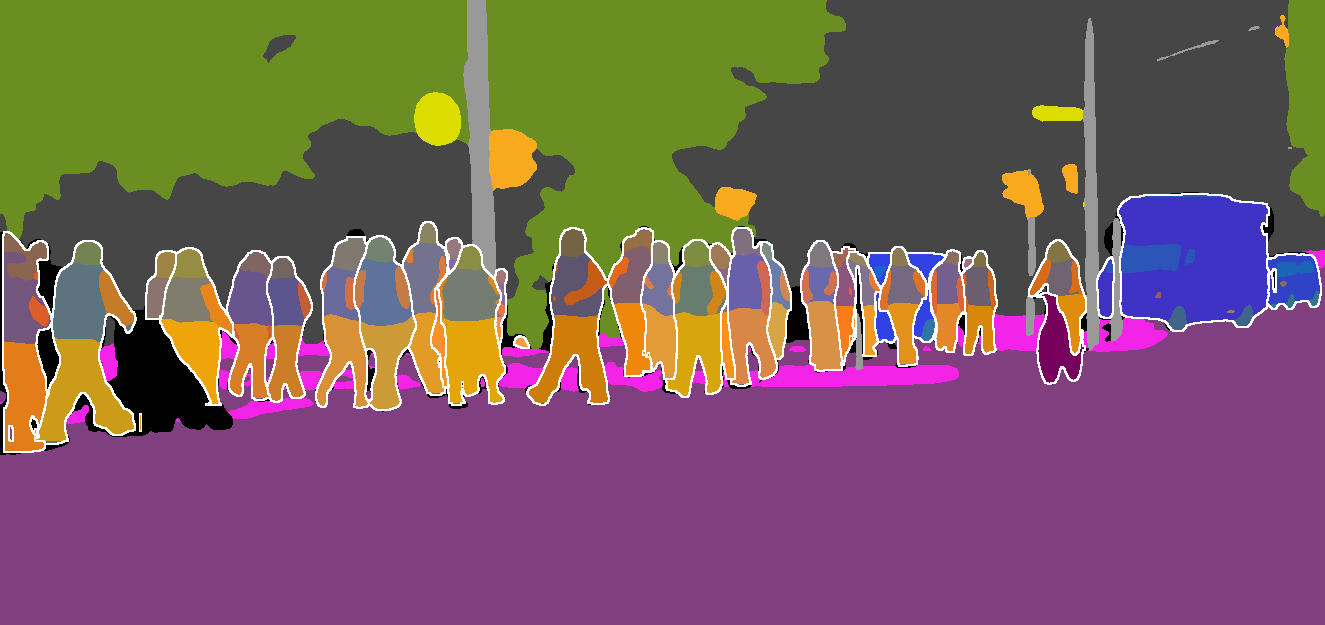}
    \end{subfigure}
    
    \vspace*{\fill}

        \begin{subfigure}{0.33\linewidth}
\includegraphics[width=\linewidth]{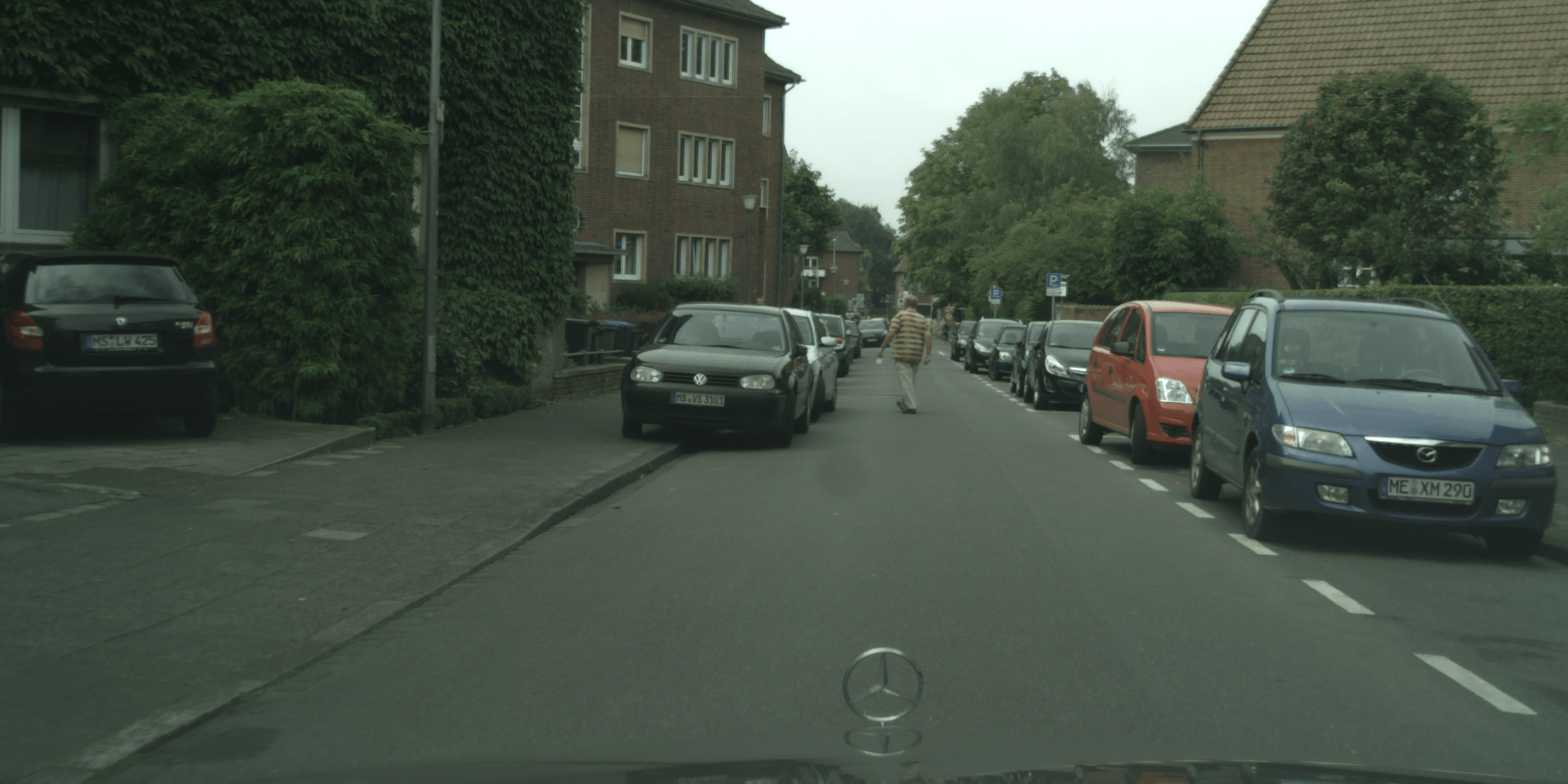}
    \end{subfigure}\hspace*{\fill}
     \begin{subfigure}{0.33\linewidth}
\includegraphics[width=\linewidth]{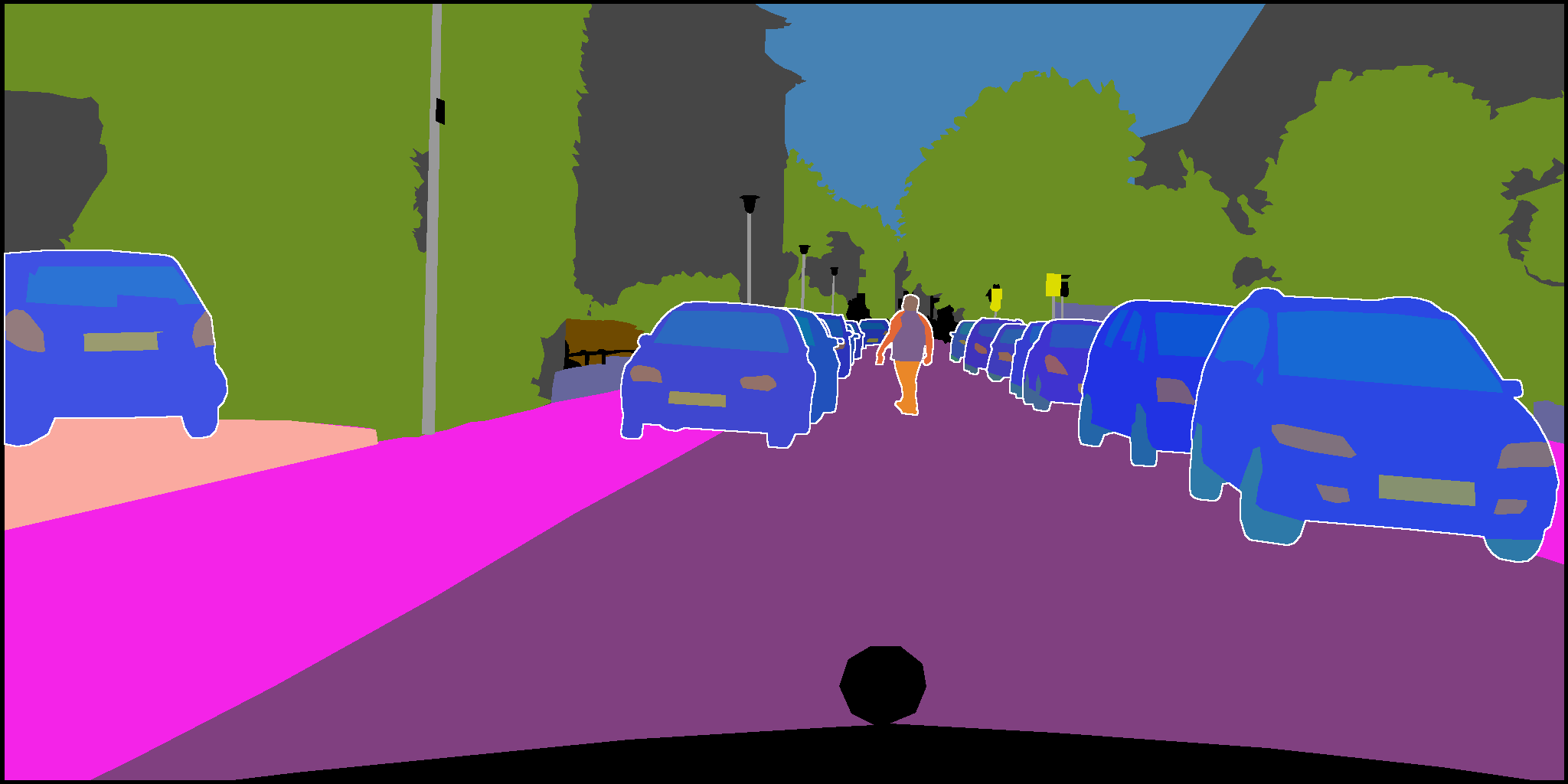}
    \end{subfigure}\hspace*{\fill}
     \begin{subfigure}{0.33\linewidth}
\includegraphics[width=\linewidth]{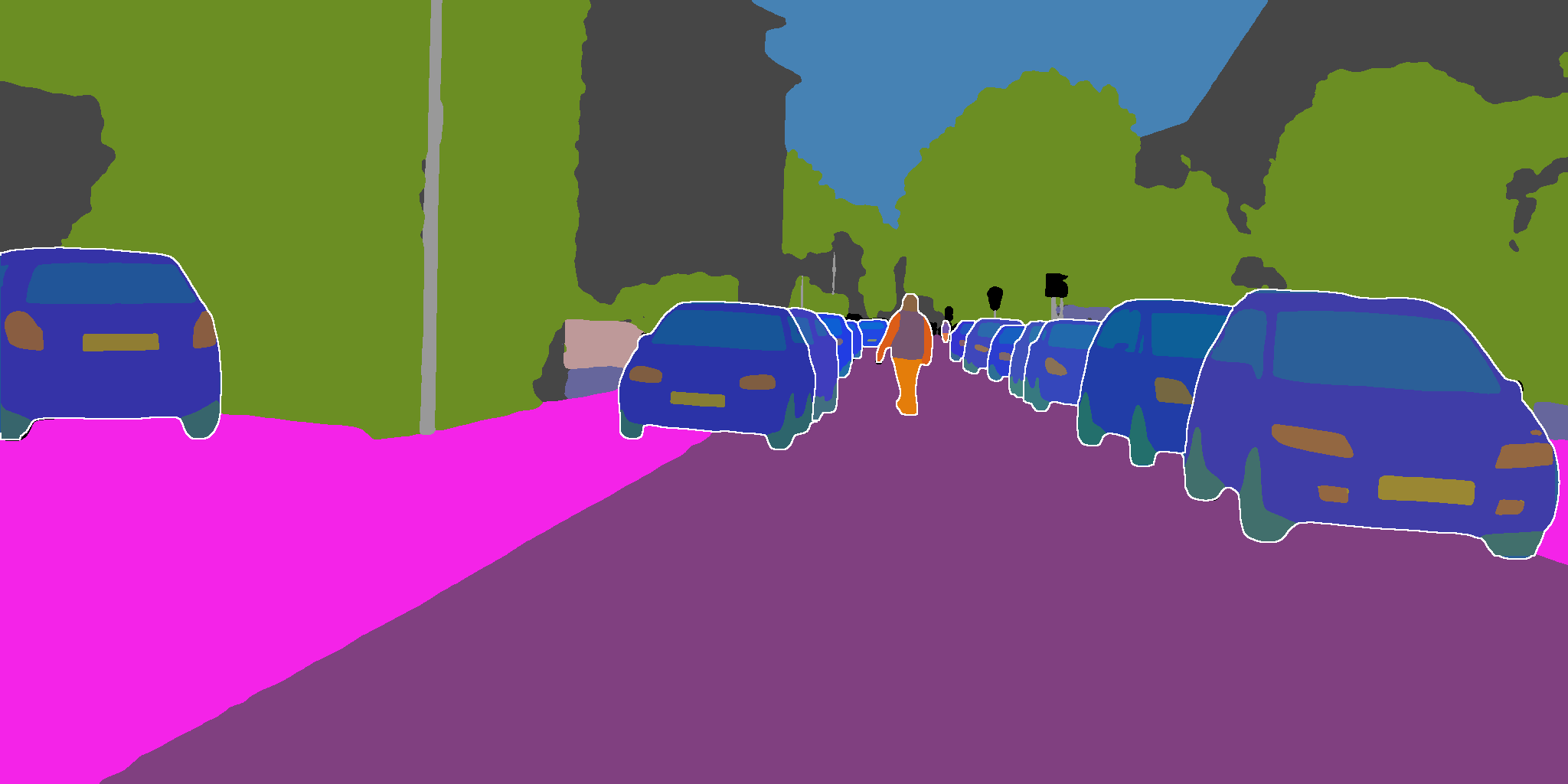}
    \end{subfigure}
    
    \vspace*{\fill}
    
        \begin{subfigure}{0.33\linewidth}
\includegraphics[width=\linewidth]{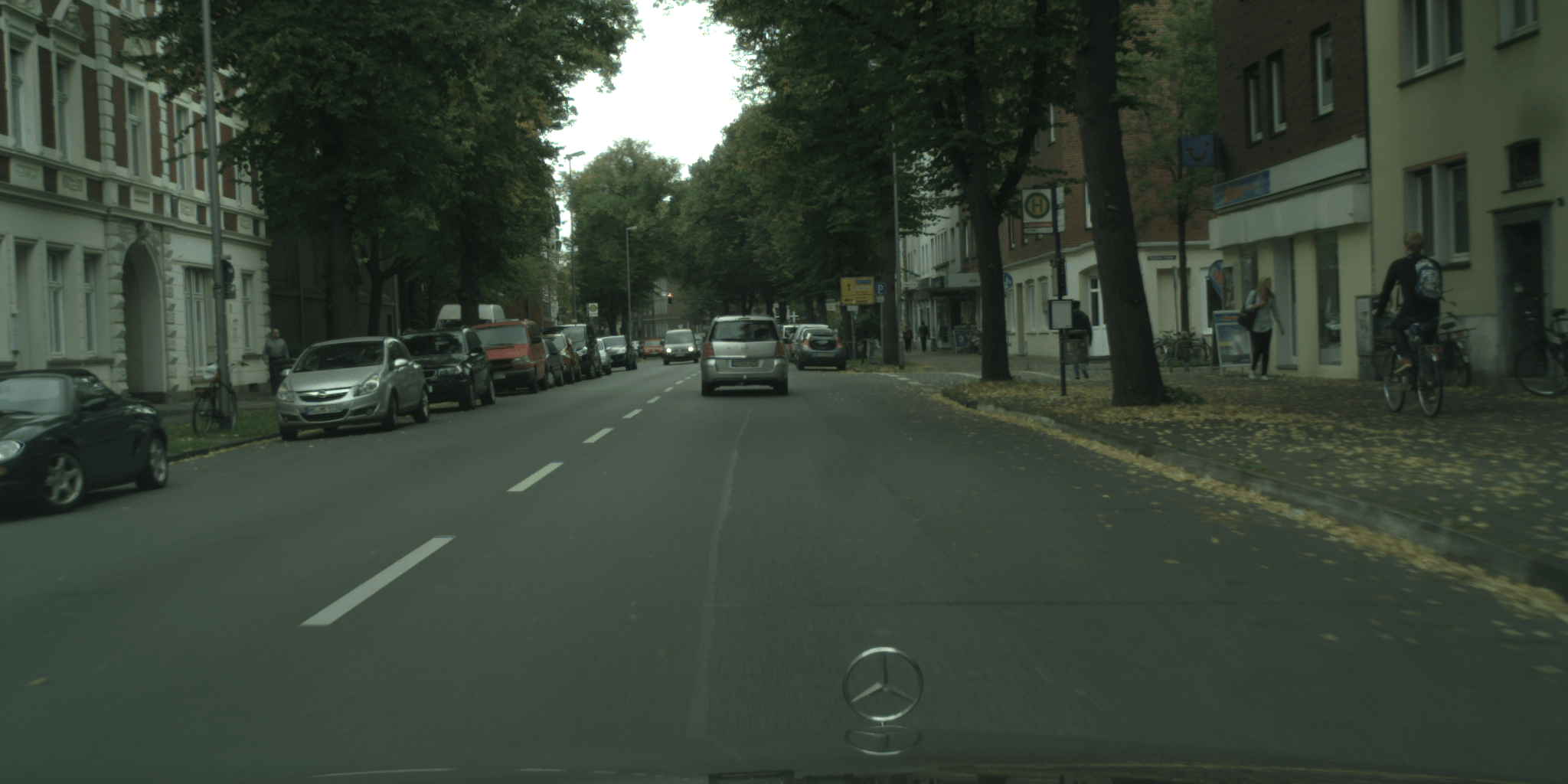}
    \end{subfigure}\hspace*{\fill}
     \begin{subfigure}{0.33\linewidth}
\includegraphics[width=\linewidth]{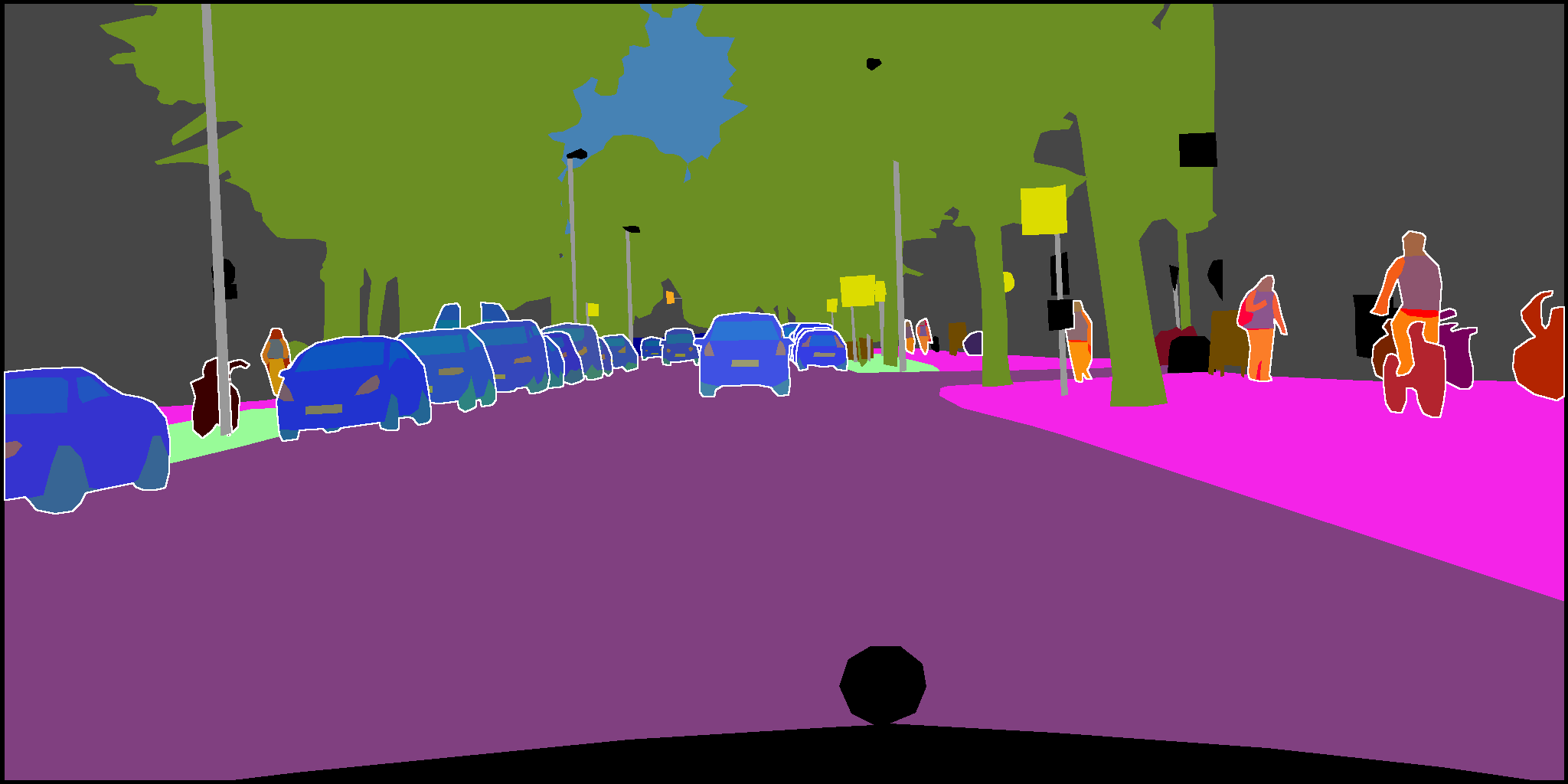}
    \end{subfigure}\hspace*{\fill}
     \begin{subfigure}{0.33\linewidth}
\includegraphics[width=\linewidth]{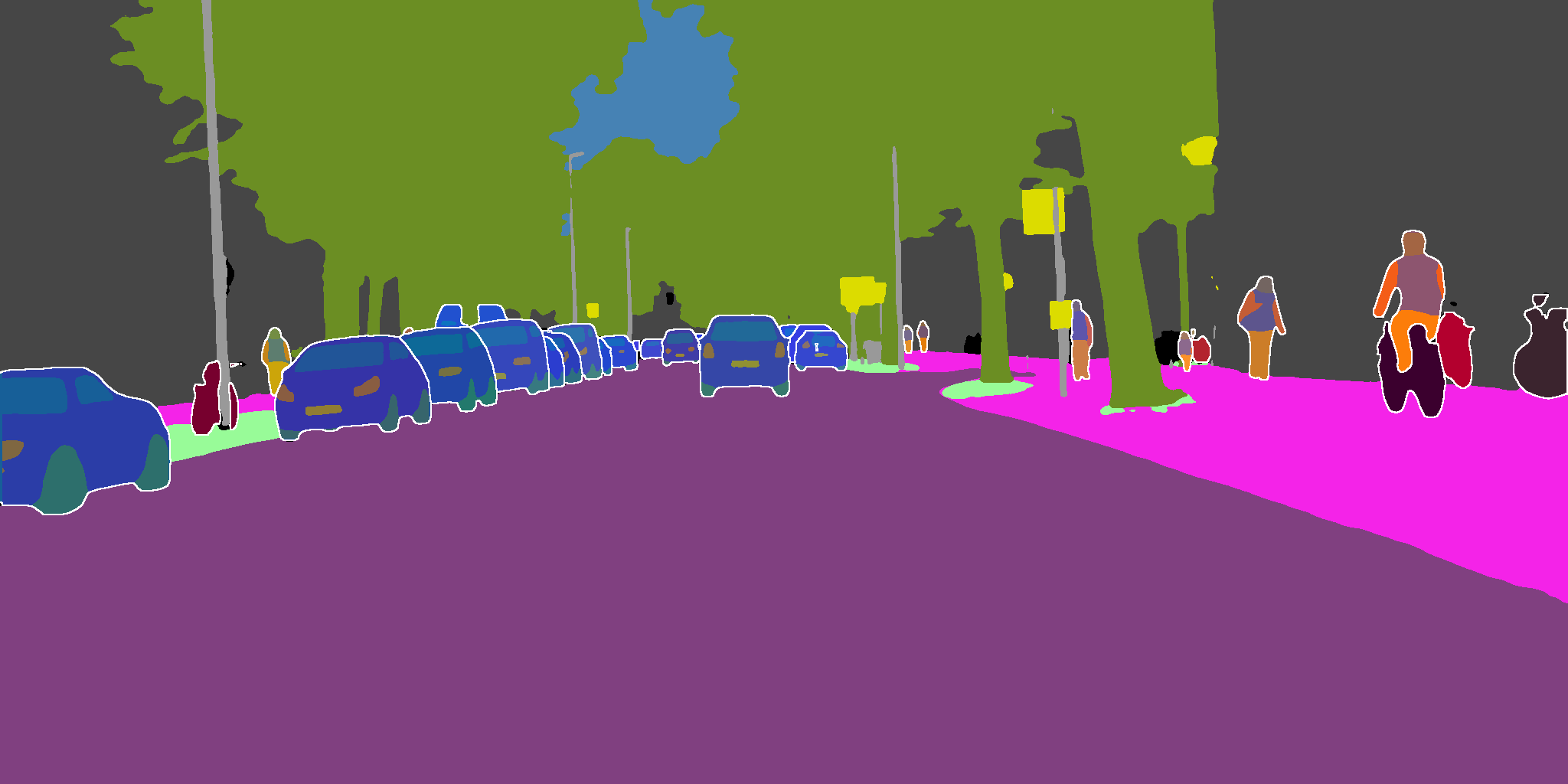}
    \end{subfigure}
    
    \vspace*{\fill}

        \begin{subfigure}{0.33\linewidth}
\includegraphics[width=\linewidth]{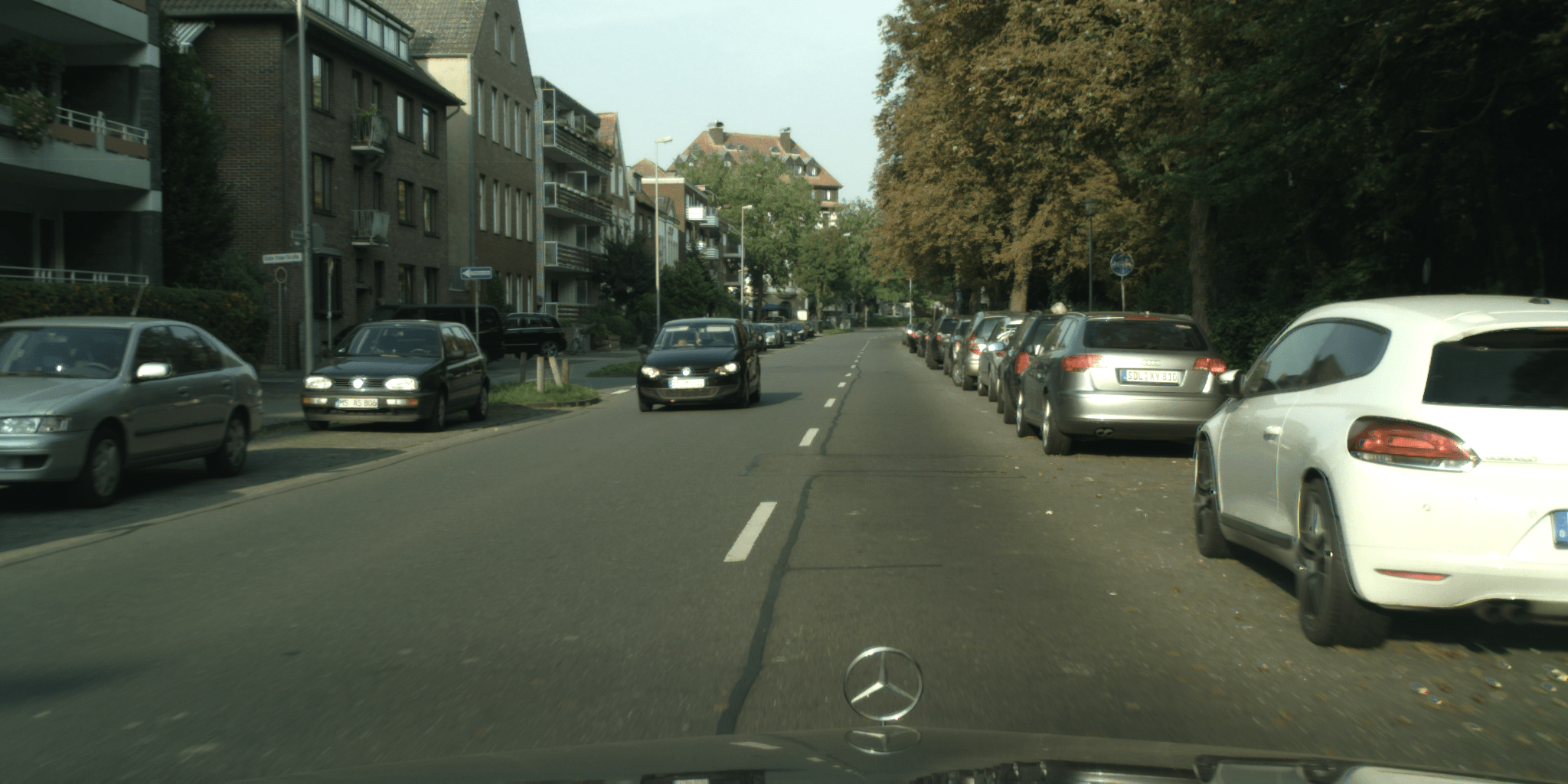}
\subcaption*{Original Image}
    \end{subfigure}\hspace*{\fill}
     \begin{subfigure}{0.33\linewidth}
\includegraphics[width=\linewidth]{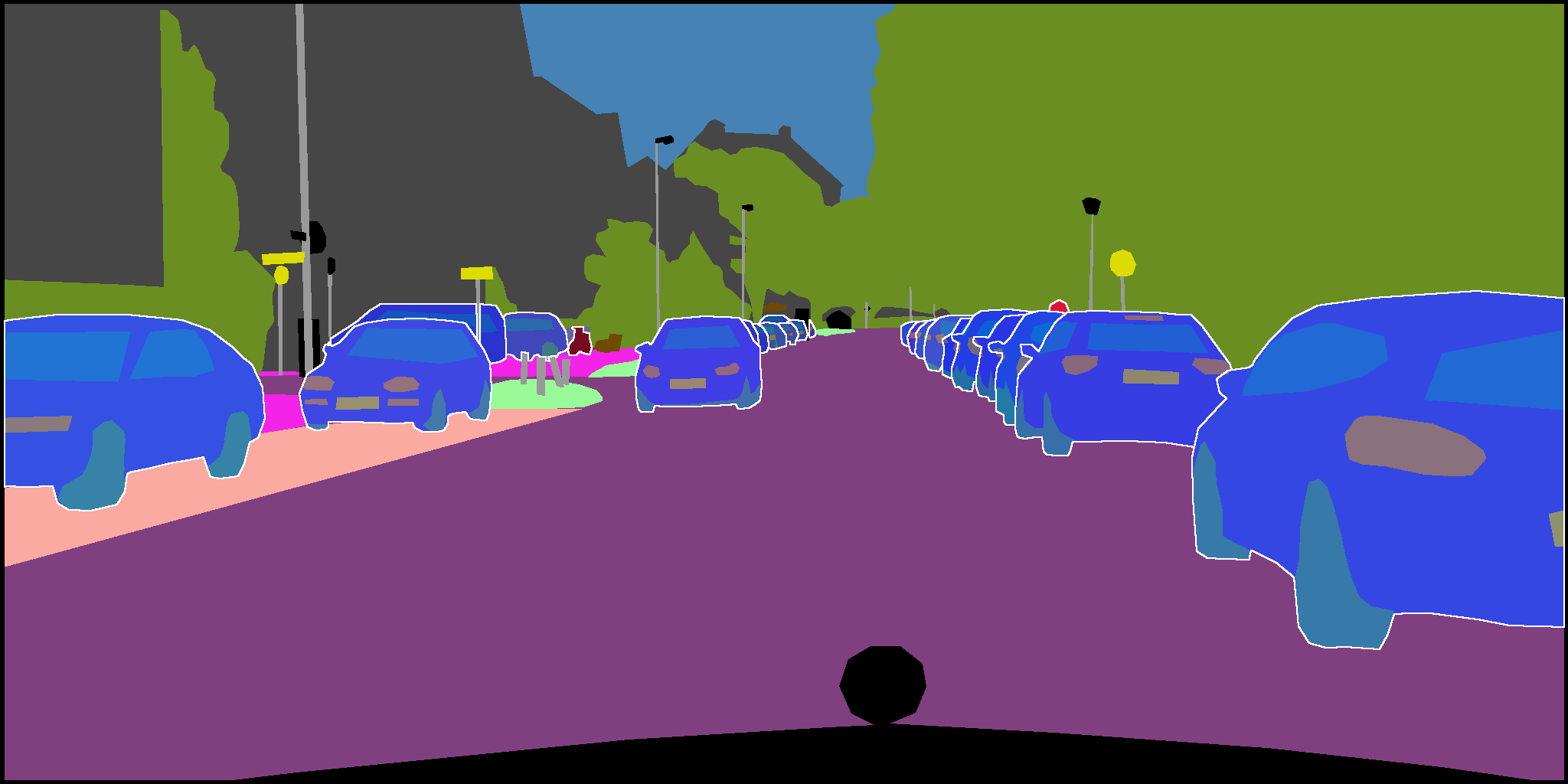}
\subcaption*{Ground-truth}
    \end{subfigure}\hspace*{\fill}
     \begin{subfigure}{0.33\linewidth}
\includegraphics[width=\linewidth]{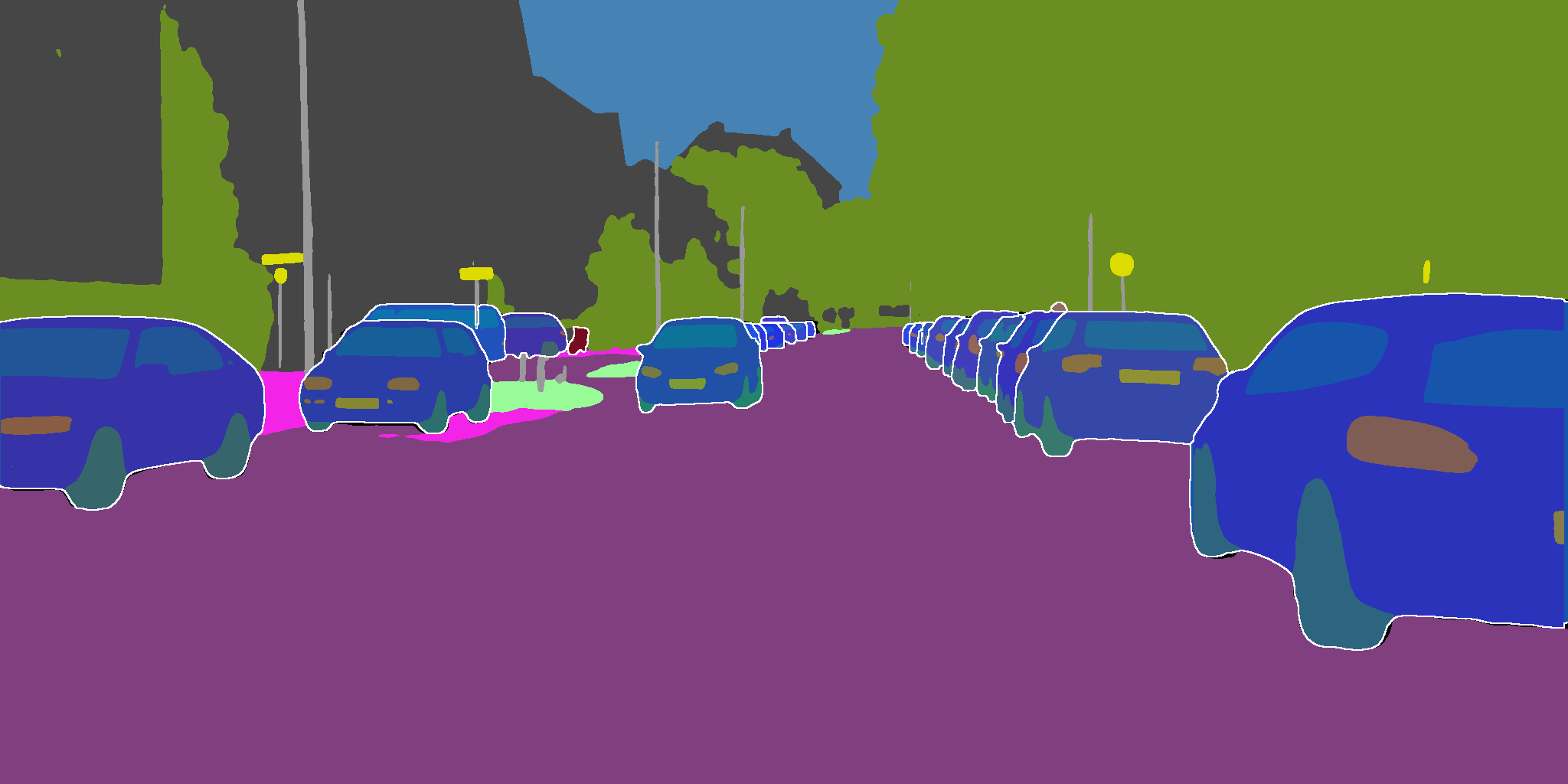}
\subcaption*{\ours}
    \end{subfigure}

    \caption{Qualitative results of our proposed model compared to the ground truth and the reference image on CPP \citep{meletis2020cityscapes}.}
    \label{fig:visualcityscapes}
\end{figure*}

\begin{figure*}[p]
    \centering
    \begin{subfigure}{0.33\linewidth}
\includegraphics[width=\linewidth]{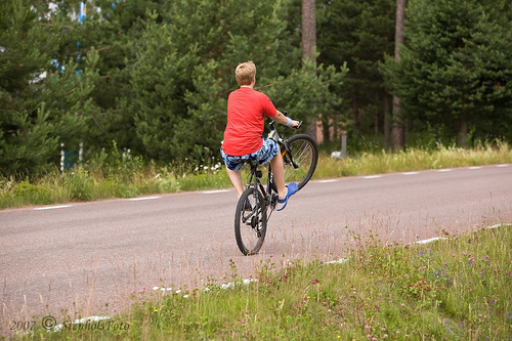}
    \end{subfigure}\hspace*{\fill}
     \begin{subfigure}{0.33\linewidth}
\includegraphics[width=\linewidth]{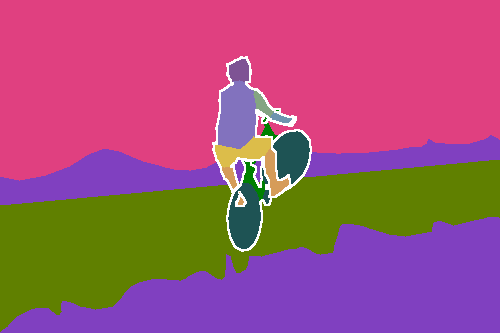}
    \end{subfigure}\hspace*{\fill}
     \begin{subfigure}{0.33\linewidth}
\includegraphics[width=\linewidth]{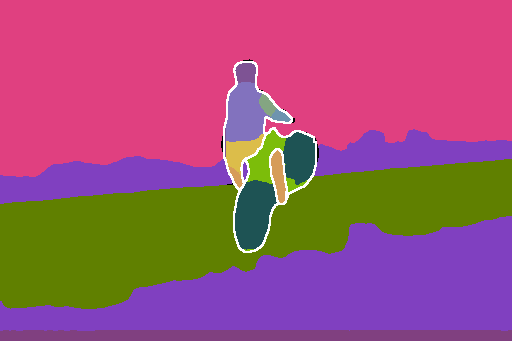}
    \end{subfigure}
    
    \vspace*{\fill}

        \begin{subfigure}{0.33\linewidth}
\includegraphics[width=\linewidth]{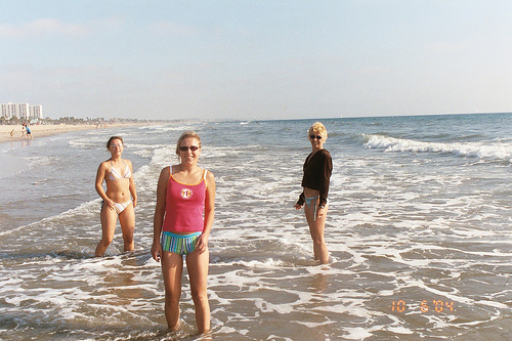}
    \end{subfigure}\hspace*{\fill}
     \begin{subfigure}{0.33\linewidth}
\includegraphics[width=\linewidth]{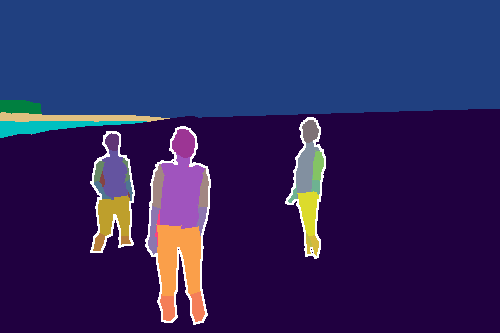}
    \end{subfigure}\hspace*{\fill}
     \begin{subfigure}{0.33\linewidth}
\includegraphics[width=\linewidth]{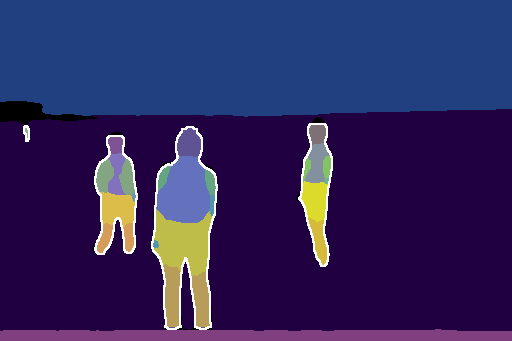}
    \end{subfigure}
    
    \vspace*{\fill}

        \begin{subfigure}{0.33\linewidth}
\includegraphics[width=\linewidth]{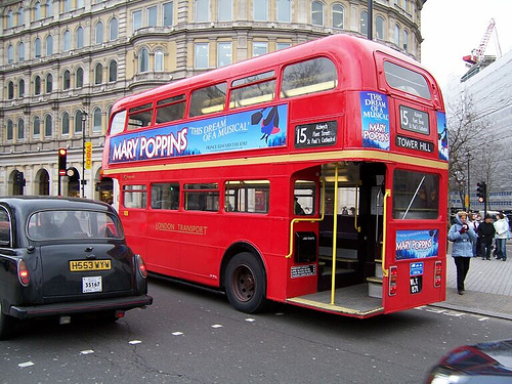}
    \end{subfigure}\hspace*{\fill}
     \begin{subfigure}{0.33\linewidth}
\includegraphics[width=\linewidth]{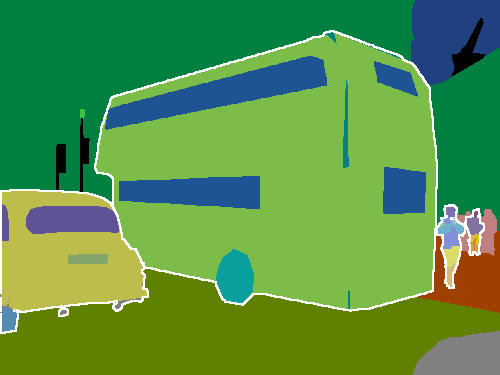}
    \end{subfigure}\hspace*{\fill}
     \begin{subfigure}{0.33\linewidth}
\includegraphics[width=\linewidth]{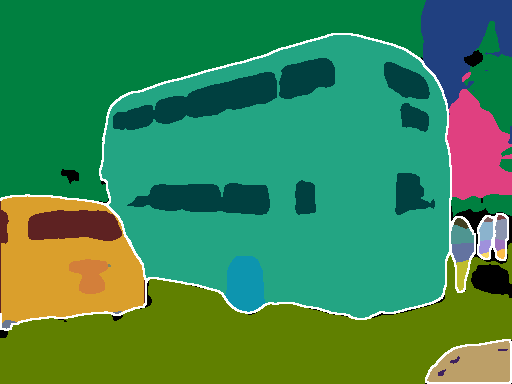}
    \end{subfigure}
    
    \vspace*{\fill}
    
        \begin{subfigure}{0.33\linewidth}
\includegraphics[width=\linewidth]{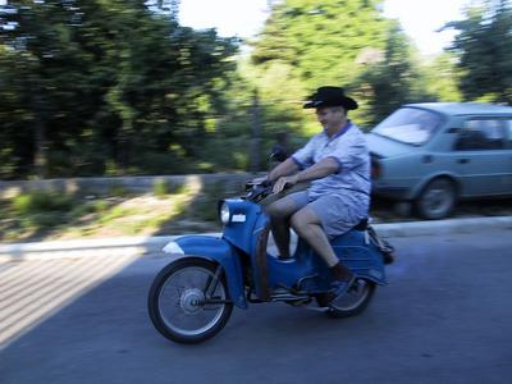}
    \end{subfigure}\hspace*{\fill}
     \begin{subfigure}{0.33\linewidth}
\includegraphics[width=\linewidth]{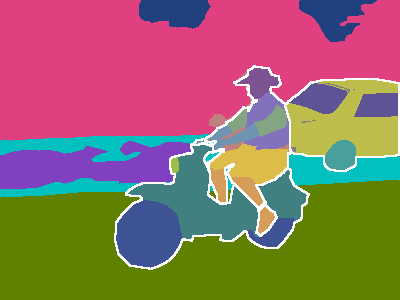}
    \end{subfigure}\hspace*{\fill}
     \begin{subfigure}{0.33\linewidth}
\includegraphics[width=\linewidth]{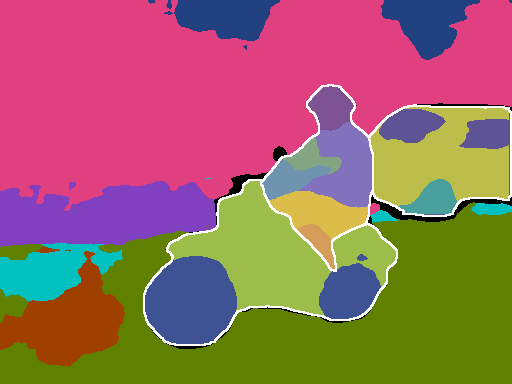}
    \end{subfigure}
    
    \vspace*{\fill}

        \begin{subfigure}{0.33\linewidth}
\includegraphics[width=\linewidth]{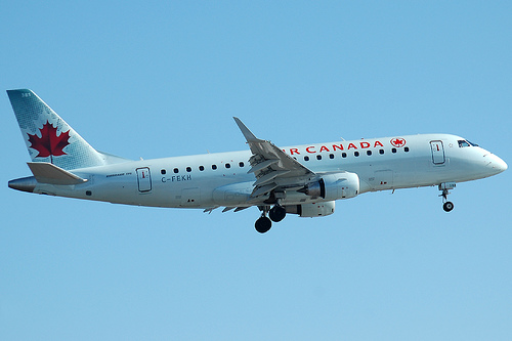}
\subcaption*{Original Image}
    \end{subfigure}\hspace*{\fill}
     \begin{subfigure}{0.33\linewidth}
\includegraphics[width=\linewidth]{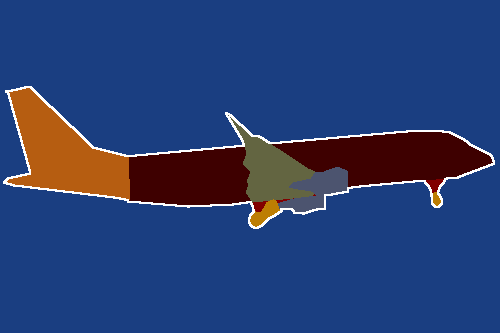}
\subcaption*{Ground-truth}
    \end{subfigure}\hspace*{\fill}
     \begin{subfigure}{0.33\linewidth}
\includegraphics[width=\linewidth]{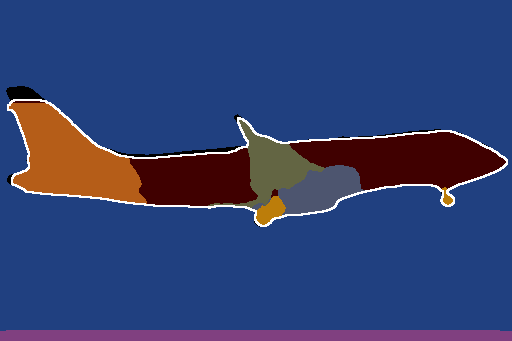}
\subcaption*{\ours}
    \end{subfigure}

    \caption{Qualitative results of our proposed model compared to the ground truth and the reference image on PPP \citep{meletis2020cityscapes}.}
    \label{fig:visualpascal}
\end{figure*}

\end{document}